\documentclass[10pt,twocolumn,letterpaper,final]{article}

\usepackage[pagenumbers]{iccv} 
\definecolor{iccvblue}{rgb}{0.21,0.49,0.74}
\usepackage[pagebackref,breaklinks,colorlinks,allcolors=iccvblue]{hyperref}
\usepackage{multirow}
\usepackage{float}
\def\paperID{7847} 
\def\confName{ICCV}
\def\confYear{2025}
\usepackage{graphicx}
\usepackage[export]{adjustbox}  
\usepackage{float}
\usepackage{indentfirst}
\usepackage{tabularx}
\title{GM-MoE: Low-Light Enhancement with Gated-Mechanism\\Mixture-of-Experts}

\author{
Minwen Liao$^{1\thanks{These three authors contributed equally to this work and are considered co-first authors.}}$ \quad
Haobo Dong$^{2\footnotemark[1]}$ \quad
Xinyi Wang$^{3\footnotemark[1]}$ \quad
Kurban Ubul$^{1\thanks{Corresponding authors.}}$ \\
Yihua Shao$^{4\footnotemark[2]}$ \quad
Ziyang Yan$^{5\footnotemark[2]}$ \\
$^1$ Xinjiang University \quad
$^2$Harbin University of Commerce \\
$^3$Changchun University of Science and Technology \\
$^4$University of Science and Technology Beijing \quad
$^5$University of Trento
}

\begin{document}
\maketitle
 \begin{abstract}
Low-light enhancement has wide applications in autonomous driving, 3D reconstruction, remote sensing, surveillance, and so on, which can significantly improve information utilization. However, most existing methods lack generalization and are limited to specific tasks such as image recovery. To address these issues, we propose \textbf{Gated-Mechanism Mixture-of-Experts (GM-MoE)}, the first framework to introduce a mixture-of-experts network for low-light image enhancement. GM-MoE comprises a dynamic gated weight conditioning network and three sub-expert networks, each specializing in a distinct enhancement task. Combining a 
self-designed gated mechanism that dynamically adjusts the weights of the sub-expert networks for different data domains. Additionally, we integrate local and global feature fusion within sub-expert networks to enhance image quality by capturing multi-scale features. Experimental results demonstrate that the GM-MoE achieves superior generalization compared to over 20 existing approaches, reaching state-of-the-art performance on PSNR on 5 benchmarks and SSIM on 4 benchmarks, respectively. Code is available at:
\url{https://github.com/Sameenok/gm-moe-lowlight-enhancement.git}
\end{abstract}    
\section{Introduction}
\label{sec:intro}
\begin{figure}[ht] 
\centering 
\includegraphics[width=8.5cm]{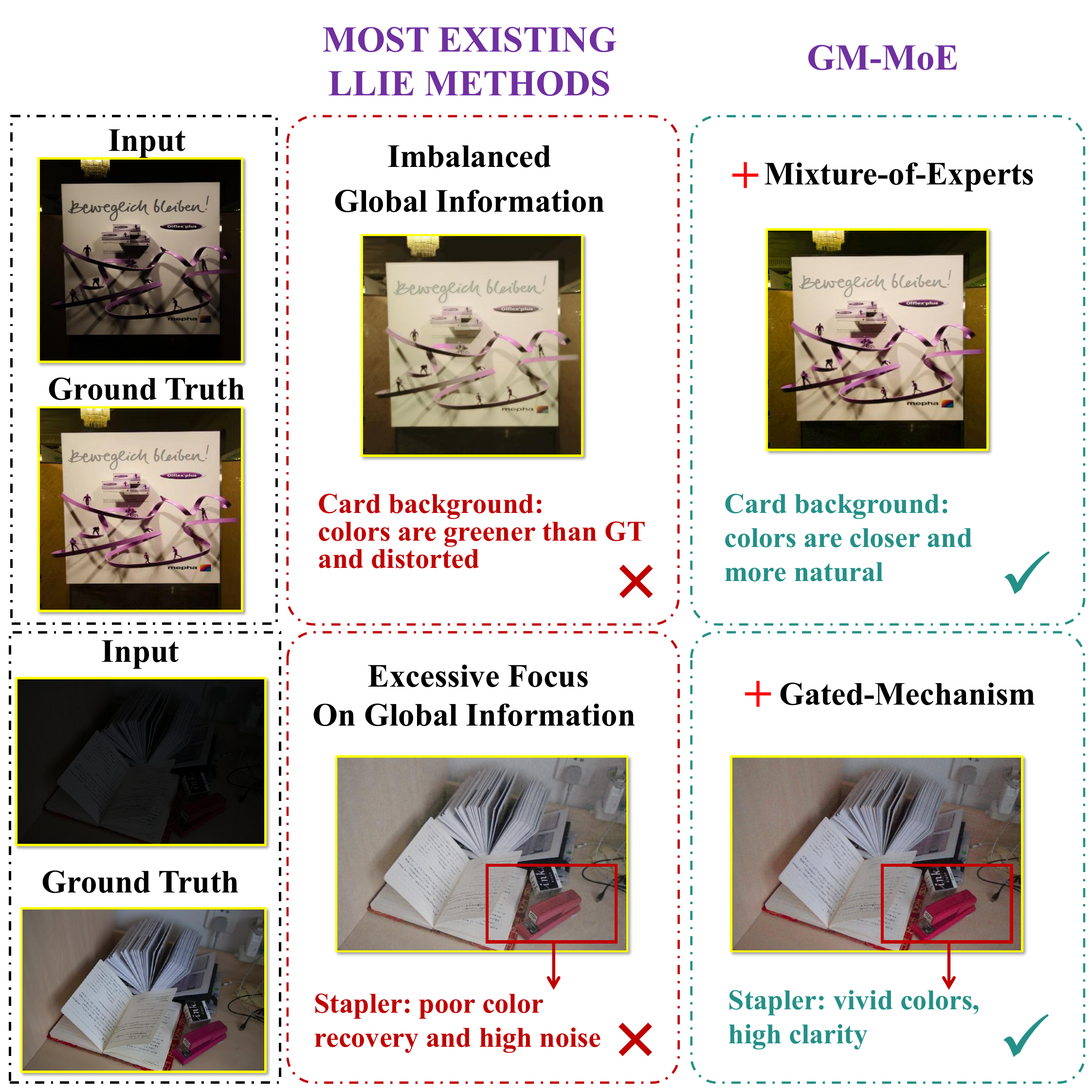} 
\caption{\small Given a low-light image, our GM-MoE achieves better performance (for both object and whole scene) compared with LightenDiffusion~\cite{jiang2024lightendiffusionunsupervisedlowlightimage}.}
\label{net2} 
\vspace{-1.0em}
\end{figure}

Low-light image enhancement (LLIE) is a crucial research area with diverse applications, including autonomous driving \cite{yan2024renderworld}, low-light scene reconstruction \cite{remondino2023critical, yan20243dsceneeditor}, remote sensing \cite{remondino2023critical,yao2024spatial,singh2022low}, and image/video analysis \cite{shao2024accidentblip, shao2025context, shao2024gwq}, since it enhances visibility and preserves fine details, enabling more reliable scene understanding in challenging lighting conditions. Although recent advancements~\cite{ref46,ref47} have improved LLIE performance, most existing methods focus on addressing specific challenges, such as noise suppression or detail restoration, rather than providing a comprehensive solution for diverse low-light scenarios. First, many existing algorithms have the problem of unbalanced global information due to local enhancement. Traditional algorithms such as histogram equalization enhance the image through a single strategy, which often over-enhances local areas, resulting in loss of image details or overexposure~\cite{gonzalez2008digital}. Convolutional Neural Networks(CNN)-based methods use a network with multiple layers of convolutions~\cite{chen2018learning,lore2017llnet}, which makes it difficult to learn the global distribution of illumination and restore it~\cite{ref16}. 

The transformer will ignore local color continuity due to an excessive focus on global information~\cite{204}, which leads to the problem of color distortion. At the same time, there is a problem of insufficient cross-domain generalization ability. Existing methods, such as SurroundNet~\cite{ref24},  are usually trained on specific datasets, and the model design lacks consideration of photos from different data domains. This results in a sharp decline in performance under unknown lighting conditions, making it difficult to achieve robust image enhancement and poor generalization ability. Meanwhile, because the problems of noise, color distortion, and blurred details in low-light images are coupled with each other, it is difficult for a single model to be optimized collaboratively. For example, suppressing noise may result in a sacrifice of details, and increasing the brightness of low-light areas may amplify color distortion. Therefore, it is a difficult problem to solve and balance the effect of image recovery. These problems limit the application of LLIE technology in complex scenes, and there is an urgent need for a unified framework that can not only enhance multiple tasks but also dynamically adapt to different lighting scenes.

\begin{figure}[ht]
    \centering
    \begin{minipage}[b]{0.48\textwidth} 
        \centering
        \includegraphics[width=\textwidth]{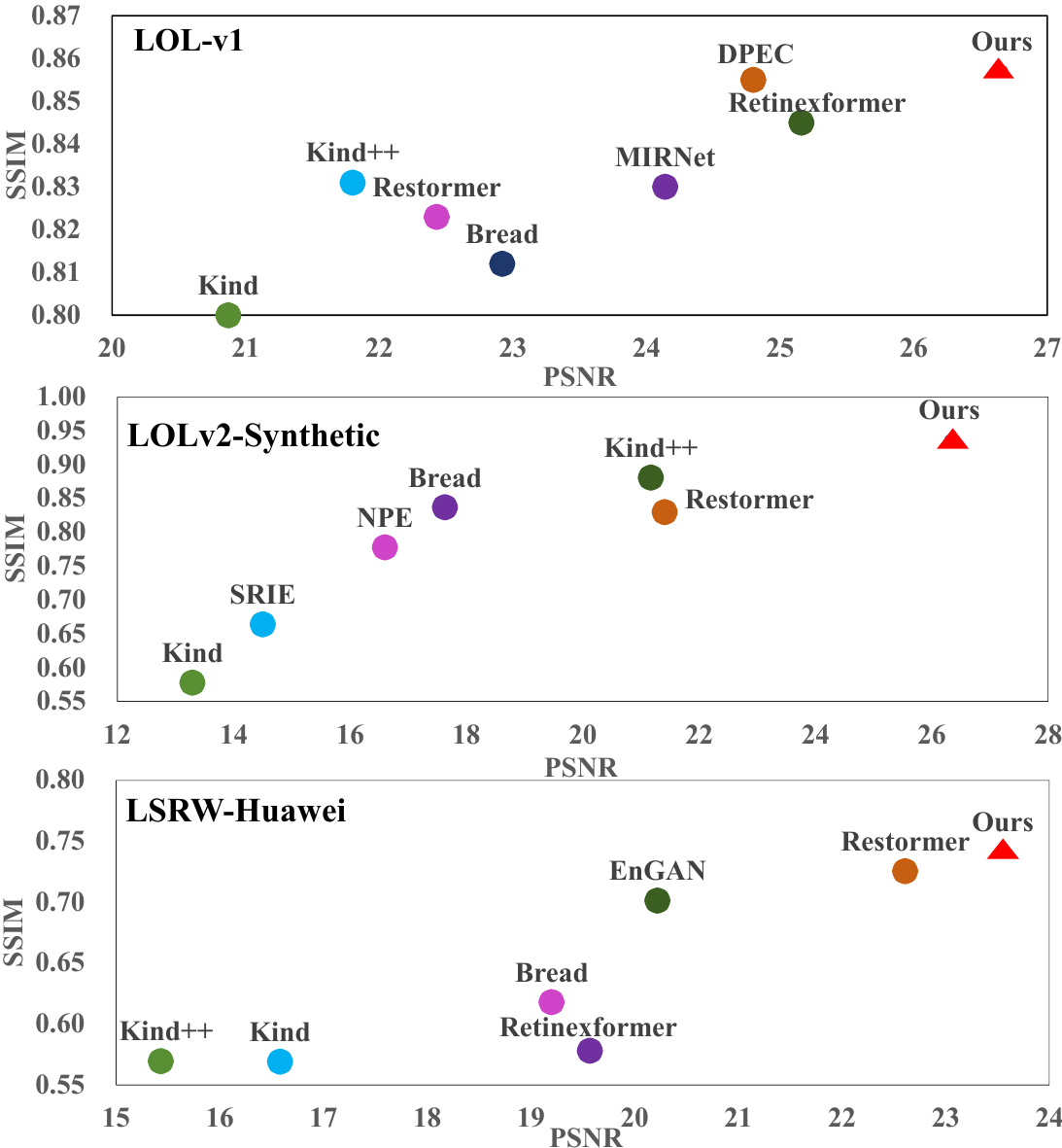}
        \caption{\textbf{The comparison results among GM-MoE and the SOTA low-light image enhancement methods on the LOL-v1 , LOLv2-Synthetic and LSRW-Huawei benchmarks.} GM-MoE outperforms all of compared approaches on both PSNR and SSIM metrics.}
        \label{comparison1}
    \end{minipage}
    \label{comparison} 
    \vspace{-2.0em}
\end{figure}

To address these issues, we propose an innovative \textbf{Gated-Mechannism Mixture-of-Experts (GM-MoE)} system for low-light image enhancement. The method is based on an improved U-Net \cite{ronneberger2015u} architecture, incorporating a gated mechanism expert network with dynamic weight adjustment to adapt to photo inputs from different data domains. GM-MoE consists of three sub-expert networks, each of which is used to solve different image enhancement tasks, namely color correction, detail recovery, and problems. The gating mechanism assigns appropriate weights to each sub-expert based on the different lighting and scene conditions of the image, achieving a balance between the image enhancement problems and thus achieving the best image enhancement under different lighting conditions and scenes. As shown in Fig. \ref{comparison1}, our framework achieves a higher PSNR and SSIM in multiple data sets compared to other state-of-the-art methods discussed in the literature.

In summary, our contributions are as follows:
\begin{itemize}

\item  We are the first to apply GM-MoE to low-light image enhancement, proposing a method that combines a dynamic gating weight adjustment network with a multi-expert module to achieve effective generalization across different data domains.
\item We propose a dynamic gating mechanism that adaptively adjusts the MoE weights according to varying lighting conditions, thereby optimizing image enhancement.
\item Our model achieves superior performance across multiple datasets and downstream tasks. Extensive experiments demonstrate that GM-MoE surpasses SOTA methods on different metrics across five datasets while maintaining strong generalization ability.
\end{itemize}

\begin{figure*}[ht]
  \centering
  \includegraphics[width=\textwidth]{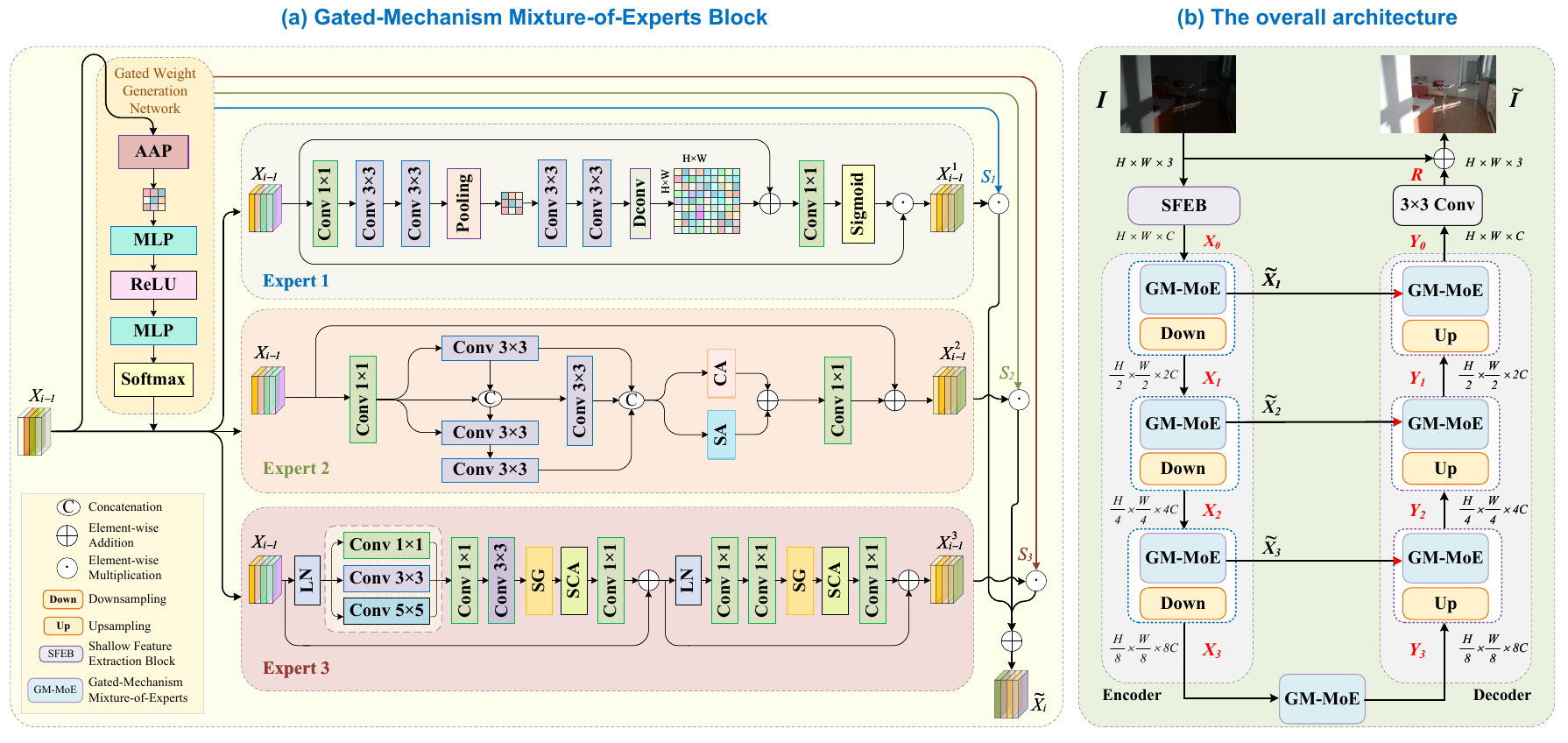}
  \caption{\textbf{Overview of the proposed GM-MOE.} (a) The GM-MoE module comprises a gated weight generation network and three specialized sub-expert networks. (b) The overall network adopts a U-Net-like encoder-decoder architecture. Given an input image, it first undergoes processing through the Shallow Feature Extraction Block (SFEB). Then, the GM-MoE module facilitates multi-scale feature fusion via multiple downsampling and upsampling operations, ultimately generating the enhanced output image.
}
  \label{fig:yourlabel}
  \vspace{-1em}
\end{figure*}
\section{Related Work} 
\label{sec:Related Work}
\subsection{Low-light Image Enhancement}
\noindent\textbf{Intensity Transformation Techniques.} Traditional low-light enhancement methods improve low-light images by directly processing pixel intensity values, including histogram equalization (HE) \cite{gonzalez2008digital,ref19} and Gamma Correction (GC) \cite{ref5,ref6,ref7}. HE enhances the contrast by redistributing the intensity histogram of the image, but it tends to over-enhance and, therefore, often amplifies noise. On the other hand, GC adjusts brightness using a nonlinear transformation, but it does not adapt well to complex lighting conditions, which leads to unnatural visual effects. In addition, adaptive contrast enhancement methods \cite{ref8,ref9} modify contrast based on local pixel statistics to enhance details, but they may inadvertently introduce noise. These methods often fail to take into account the full complexity of lighting, leading to perceptual differences compared to images captured under typical lighting conditions.

\noindent\textbf{Perception-Based Models.} To compensate for these deficiencies, some methods simulate the human visual mechanism, such as the Retinex theory, which decomposes an image into reflection and illumination components. Multi-scale Retinex (MSR) \cite{ref13,ref14} enhances contrast at different scales but may lead to color differences due to inaccurate light estimation. The dark channel prior (DCP) \cite{ref15,ref16}, originally used for defogging, was later adapted for low-light enhancement but is prone to color oversaturation in complex scenes. Recently, the global brightness ranking regularization method proposed by Li et al., \cite{ref17} has improved visual effects, though it can still introduce issues during enhancement.

\noindent\textbf{Deep Learning-Based Approaches.} In recent years, deep learning has promoted the application of convolutional neural networks in low-light image enhancement \cite{7782813,ref20,ref21,ref22,ref23,ref24,ref25,204,ref28}. For example, Chen et al., \cite{chen2018learning} proposed the SID model to directly convert low-light images to normal-light images; Guo et al.,\cite{ref19} used a lightweight network to implement pixel-level curve estimation (Zero-DCE); Jiang et al.,\cite{ref32} employed GANs for unsupervised learning in EnlightenGAN. In addition, Wei et al. proposed Retinex-Net \cite{wei2018deep}, Zhang et al., introduced KinD \cite{ref34}, and Liu et al., improved the RUAS model \cite{ref25}, all of which significantly enhanced image restoration performance. In the field of Transformers, Liang et al., \cite{ref25} proposed SwinIR, and Zamir et al., \cite{204} developed Restormer, both of which achieve image restoration by capturing global features. Meanwhile, Wang et al., \cite{ref24} introduced LLFlow, and Xu et al., \cite{ref39} proposed the SNR-Aware method, utilizing normalizing flow and signal-to-noise ratio optimization, respectively, to enhance image details. Although these methods have made significant progress in low-light image enhancement, most of them adopt a single neural network structure. When dealing with different lighting conditions and complex scenes, their generalization ability remains limited, making it difficult to simultaneously perform multiple tasks such as noise suppression, detail restoration, and color correction across different datasets.

\subsection{Multi-Expert Systems}
MoE was originally proposed by Jacobs et al.,~\cite{ref45} and aims to build a system consisting of multiple independent networks (experts), each responsible for processing a specific subset of data. The approach emphasizes that combining the expertise of different models can significantly improve the overall performance when dealing with complex tasks. In recent years, MoE techniques have demonstrated excellence in several domains, including image recognition~\cite{ref47}, machine translation~\cite{ref51}, scene parsing~\cite{ref52}, speech recognition~\cite{ref53}, and recommender systems~\cite{ref54}. 

In the field of low-light image enhancement, there is a lack of MoE applications. The expert network module possesses the capability to simultaneously address multiple low-light image enhancement challenges, a characteristic that grants it unique advantages in the field of low-light image enhancement. Moreover, existing methods exhibit significant limitations: most low-light enhancement models can only achieve satisfactory restoration results for a single problem and lack the ability to adaptively adjust processing strategies based on the illumination conditions of the input image.

To address the limitations of existing methods and leverage the unique strengths of expert networks, this paper proposes a novel Multi-Expert Low-Light Enhancement Network. As illustrated in Fig.~\ref{fig:yourlabel}, we design a dynamically weighted multi-expert network based on the U-Net architecture, where three specialized sub-expert modules are optimized for color distortion, detail loss, and low-contrast issues, respectively. To overcome the insufficient generalization ability of existing models, we innovatively introduce a Dynamic Weight Adjustment Network, which automatically adjusts the weight allocation of the three sub-networks based on input image features, thereby achieving adaptive enhancement for images from different data domains.
\section{Method}
\label{method}

In the model we designed, the overall architecture is as follows: The input dark image 
\( I \in \mathbb{R}^{H \times W \times 3} \) 
is first processed by SFEB to obtain low-level features 
\( X_0 \in \mathbb{R}^{H \times W \times C} \),
where \( H \) and \( W \) denote the spatial dimensions of the image and \( C \) denotes the number of channels. These features are then input into a module similar to the U-Net~\cite{ronneberger2015u} network architecture, which contains encoder layers to further extract deeper features \( F_d \in \mathbb{R}^{H \times W \times 2C} \). The GM-MOE module is introduced in the encoder and decoder of each layer. During encoding, the encoder of each layer compresses the image features by gradually reducing the spatial dimension and increasing the channel capacity, while the decoder gradually restores the image resolution by upsampling the low-resolution feature map \( F_l \in \mathbb{R}^{\frac{H}{8} \times \frac{W}{8} \times 8C} \) to progressively recover the image resolution. To optimize feature recovery, pixel-shuffle techniques are introduced to improve the effects of upsampling. To aid low-light image feature recovery, the initial features are preserved and fused between the encoder and decoder via skip connections. Each layer of the GM-MOE module is responsible for fusing the lower-level features of the encoder with the higher-level features of the decoder, thereby enriching the structure and texture details of the image.
In the final stage, the deep features 
\( F_d \) further enhances the detailed features at the spatial resolution, and the residual image generated by the convolution operation \( R \) is added to the input image \( I \) to obtain the final enhanced image \( \hat{I} = I + R \).

\subsection{Gated-Mechanism Mixture-of-Experts}
GM-MOE block consists of a Gated Weight Generation Network and an expert network module, where the expert network includes a color restoration submodule, a detail enhancement submodule, and an advanced feature enhancement network module. The following sections describe these modules in turn.

In order to achieve adaptive feature extraction in different data domains, we propose a GM-MoE network. First, the input image is passed through adaptive average pooling to convert the image features into a feature vector. Then, this feature vector passes through a fully connected layer with an activation function, and then through another fully connected layer to project onto three expert networks to generate the weights \(s_1, s_2, s_3\). These weights enable the network to dynamically adjust its parameters based on photos from different data domains (i.e., different scenes and different lighting characteristics), ensuring that the sum of the weights is 1.
\begin{equation}
{S}=\left [ s_{\mathrm{1}}, s_{\mathrm{2}},s_{\mathrm{3}} \right ] ,s_{\mathrm{1}}+  s_{\mathrm{2}}+  s_{\mathrm{3}}=\mathrm{1}.
\end{equation}
Each expert network \(\text{Net}_i\) processes the input feature \(X_{i-1}\) and generates the corresponding output feature \(X^i_{i-1}\), where \(i \in \{1, 2, 3\}\):
\begin{equation}
{X}_{\mathrm{i-1}}^{\mathrm{i}} ={Net}_{i} \left ( {X}_{i-1} \right ) .
\end{equation}
The final output feature \(\tilde{X}_i\) is obtained by summing the outputs of all expert networks weighted by their weights:
\begin{equation}
\tilde{X} _{i} =s_{1}{X}_{i-1}^{1}+ s_{2}{X}_{i-1}^{2}+s_{3}{X}_{i-1}^{3}.
\end{equation}
This adaptive weighting mechanism combined with multiple expert networks enables the gated weight generation network to effectively capture domain-specific features after feature extraction, thereby improving the robustness and adaptability of the model to photos in different data domains.

As shown in Figure~\ref{fig:yourlabel}, the color restoration expert network (Expert1, also named Net1) is used to restore the color information of images under low light conditions. 
The color restoration subnetwork we designed employs pooling operations to focus on key color features during downsampling while simultaneously learning image information. Additionally, deconvolution operations are utilized to precisely restore image details, with nonlinear interpolation adopted to ensure smooth and natural color transitions:
\begin{equation}
Y_{1}\left ( i,j \right )=\sum_{m,n} {w}_{mn} {X}\left ( i+m,j+n \right ) 
\end{equation}
where \(Y_{1}(i, j)\) denotes the output at position \((i,j)\), \(X(i+m,j+n)\) is the input feature at the adjacent position, and \(w_{mn}\) is the interpolation weight, which satisfies:
\(w_{mn} = 1\)
to ensure luminance consistency. To preserve the original image characteristics, the processed tensor is connected to the input through a residual connection. Finally, a Sigmoid activation function is used to limit the color output to the interval \([0,1]\), reducing color anomalies and oversaturation problems and ensuring that the enhanced image colors are natural and realistic.

As shown in Fig.~\ref{fig:yourlabel}, the detail enhancement subnetwork (Expert2, also named Net2) uses convolutions and attention mechanisms to enhance image details. The network uses different attention mechanisms in combination for feature extraction. Among them, important channel features are extracted through the channel attention mechanism. At the same time, the spatial attention mechanism is used, which combines Max Pooling and Avg Pooling, and then processed by convolution.
To fuse the characteristics of different attention mechanisms, we concatenate different attentions, where max pooling and average pooling are used to focus on key spatial positions in the image. Finally, the outputs of channel attention and spatial attention are combined with the original input image through a residual connection to preserve the original features and enhance the detail recovery ability. This structure improves the detail recovery ability of the image.

As shown in Fig.~\ref{fig:yourlabel}, the advanced feature enhancement subnetwork (Expert3, also named Net3) improves image quality through convolution, multi-scale feature extraction, a gating mechanism, and an attention mechanism. The input image is passed through a multi-scale convolution to extract and fuse features. These fused features are then processed further by a gating network (SG) and a channel attention mechanism (SCA).
Finally, the enhanced features are added back to the input image via a residual connection to preserve the original details.

This method can effectively adapt to low-light scenes and improve image quality by dynamically adjusting the weights of the expert network. 

\subsection{Shallow Feature Extraction Module}

\begin{figure}[ht] 
\centering 
\includegraphics[width=\columnwidth]{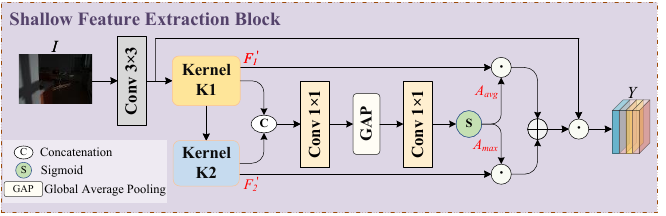} 
\caption{\textbf{Shallow Feature Extraction Block.} The architecture of SFEB uses parallel convolutional and GAP for multi-scale feature capture.}.
\label{SFEB} 
\vspace{-1.0em}
\end{figure}

To improve the effectiveness of image feature extraction and suppress invalid features, we design a multi-scale feature enhancement module (SFEB, Shallow Feature Enhancement Block) to process the input feature map \(X \in \mathbb{R}^{C \times H \times W}\). The SFEB generates a feature map \(F_1 \in \mathbb{R}^{C \times H \times W}\) through a \(3 \times 3\) depth separable convolution, as shown in Fig.~\ref{SFEB} . In addition, SFEB also obtains a feature map \(F_2 \in \mathbb{R}^{C \times H \times W}\) using different sizes of convolution kernels through hole convolution (dilation rates) to capture multi-scale spatial information.

SFEB uses two convolutions to compress the channel numbers of \(F_1\) and \(F_2\) to form the fused feature map \(F_e \in \mathbb{R}^{C' \times H \times W}\), which fuses different feature information.

To introduce the attention mechanism, SFEB performs global pooling on the fused feature map \(F_e\) to obtain the channel-weighted features \(A_{avg}\) and \(A_{max}\). Then, the attention map \(F_{w} \in \mathbb{R}^{C' \times H \times W}\) is generated by channel concatenation and a \(7 \times 7\) convolution to enhance the features of key regions:
\begin{equation} 
F_{w}=F_{1}^{'} \odot {A}_{avg}+F_{2}^{'} \odot{A}_{max}
\end{equation}
Finally, the output feature map \(Y\) is obtained by element-wise multiplication of the input feature \(X\) and the attention map \( F_{w}\):
\begin{equation} 
{Y}={X}\odot F_{w}
\end{equation}

This design, which combines a multi-scale convolution with an attention mechanism, gives SFEB stronger feature extraction capabilities, thereby improving the brightness and detail recovery of the image.

\subsection{Loss Function}
To ensure that the network-generated image \( \hat{I} \) is as close as possible to the reference image \( I_\mathrm{gt} \), we introduce the peak signal-to-noise ratio loss (PSNRLoss) as a loss function in the training to measure and maximize the quality of the output image. We first define the mean squared error (MSE) as:
\begin{equation} 
\mathrm{MSE}=\frac{1}{N}\sum_{i=1}^{N}\left ( \hat{I}\left ( i \right ) -I_{\mathrm{gt}}\left ( i \right ) \right )^{\mathrm{2}}    
\end{equation}
and then define the PSNR loss based on the MSE:
\begin{equation} 
\mathrm{PSNR\ loss} = -\frac{10}{\log{\left ( 10 \right ) } } \cdot \log{\left ( \mathrm{MSE} + \epsilon \right ) }  
\end{equation}

where \(N\) is the total number of pixels in the image, \(\hat{I}\left ( i \right )\) and \(I_{\mathrm{gt}}\left ( i \right )\) are the predicted and true values at pixel position \(i\), respectively, and \(\epsilon\) is a small positive number used to prevent the denominator from becoming zero. During training, the network weights are updated by minimizing the PSNR Loss to improve the network's image restoration ability.

\section{Experiment}
\subsection{Datasets and Implementation Details}

\begin{table*}[h!]
    \footnotesize
    \begin{center}
    \scalebox{1.0}{
    \begin{tabular}{l ccccccc}
        \toprule
        \multirow{2}{*}{Methods} & \multicolumn{2}{c}{LOL-v1~\cite{wei2018deep}} & \multicolumn{2}{c}{LOLv2-Real~\cite{9328179}} & \multicolumn{2}{c}{LOLv2-Synthetic~\cite{9328179}} & \multirow{2}{*}{\#Param (M)} \\
        \cmidrule(lr){2-3} \cmidrule(lr){4-5} \cmidrule(lr){6-7}
        & PSNR & SSIM & PSNR & SSIM & PSNR & SSIM & \\
        \midrule
        
        SID~\cite{Chen2019SeeingMI}           & 14.35 & 0.436 & 13.24 & 0.442 & 15.04 & 0.610 & 7.76   \\
        RF~\cite{Kosugi_Yamasaki_2020}        & 15.23 & 0.452 & 14.05 & 0.458 & 15.97 & 0.632 & 21.54  \\
        UFormer~\cite{wang2021uformergeneralushapedtransformer} & 16.36 & 0.771 & 18.82 & 0.771 & 19.66 & 0.871 & 5.29   \\
        EnGAN~\cite{Jiang2019EnlightenGANDL}  & 17.48 & 0.620 & 18.23 & 0.617 & 16.57 & 0.734 & 114.35 \\
        Restormer~\cite{204}                 & 22.43 & 0.823 & 19.94 & 0.827 & 21.41 & 0.830 & 26.13  \\
        Retinexformer~\cite{205}             & 25.16 & 0.845 & 22.80 & 0.840 & 25.67 & 0.930 & 1.61   \\
        DeepUPE~\cite{Wang_2019_CVPR}         & 14.38 & 0.446 & 13.27 & 0.452 & 15.08 & 0.623 & 1.02   \\
        LIME~\cite{7782813}                  & 16.76 & 0.560 & 15.24 & 0.419 & 16.88 & 0.757 & -      \\
        MF~\cite{fu2016fusion}               & 18.79 & 0.640 & 18.72 & 0.508 & 17.50 & 0.773 & -      \\
        NPE~\cite{wang2013naturalness}       & 16.97 & 0.589 & 17.33 & 0.452 & 16.60 & 0.778 & -      \\
        SRIE~\cite{fu2016weighted}           & 11.86 & 0.500 & 14.45 & 0.524 & 14.50 & 0.664 & \textcolor{blue}{0.86} \\
        RetinexNet~\cite{wei2018deep}        & 16.77 & 0.560 & 15.47 & 0.567 & 17.13 & 0.798 & \textcolor{red}{0.84}\\
        Kind~\cite{ref34}                    & 20.86 & 0.790 & 14.74 & 0.641 & 13.29 & 0.578 & 8.02   \\
        Kind++~\cite{zhang2021beyond}        & 21.80 & 0.831 & 20.59 & 0.829 & 21.17 & 0.881 & 8.27   \\
        MIRNet~\cite{ref1839}                & 24.14 & 0.830 & 20.02 & 0.820 & 21.94 & 0.876 & 31.76  \\
        SNR-Net~\cite{inproceedings}         & 24.61 & 0.842 & 21.48 & \textcolor{blue}{0.849} & 24.14 & 0.928 & 39.12  \\
        Bread~\cite{hu2021lowlightimageenhancementbreaking} & 22.92 & 0.812 & 20.83 & 0.821 & 17.63 & 0.837 & 2.12   \\
        DPEC~\cite{wang2024dpecdualpatherrorcompensation}   & 24.80 & \textcolor{blue}{0.855} & \textcolor{blue}{22.89} & \textcolor{red}{0.863} & \textcolor{blue}{26.19} & \textcolor{red}{0.939} & 2.58      \\
        PairLIE~\cite{fu2023learning}        & 23.53 & 0.755 & 19.89 & 0.778 & 19.07 & 0.794 & 0.33 \\
        LLFormer~\cite{ref28}        & \textcolor{blue}{25.76} & 0.823 & 20.06 & 0.792 & 24.04 & 0.909 & 24.55 \\
        QuadPrior~\cite{Wang_2024_CVPR}      & 22.85 & 0.800 & 20.59 & 0.811 & 16.11 & 0.758 & 1252.75   \\
        3DLUT\cite{zeng2020lut}  & 14.35 & 0.445 & 17.59 & 0.721 & 18.04 & 0.800 & 0.59 \\
        Sparse\cite{9328179} & 17.20 & 0.640 & 20.06 & 0.816 & 22.05 & 0.905 & 1.08 \\
        RUAS\cite{liu2021ruas}   & 18.23 & 0.720 & 18.37 & 0.723 & 16.55 & 0.652 & 1.03 \\
        DRBN~\cite{Yang_2020_CVPR}   & 20.13 & 0.830 & 20.29 & 0.831 & 23.22 & 0.927& 1.83 \\
        \midrule
        \textbf{Ours}                        & \textcolor{red}{\textbf{26.66}} & \textcolor{red}{\textbf{0.857}} & \textcolor{red}{\textbf{23.65}} & 0.806 & \textcolor{red}{\textbf{26.30}} & \textcolor{blue}{\textbf{0.937}} & 19.99  \\
        \bottomrule
    \end{tabular}%
    }
    \end{center}
     \vspace{-1.5em}
    \caption{\textbf{Quantitative Comparison on LOL-v1, LOLv2-Real, and LOLv2-Synthetic Datasets.} 
    The best results are highlighted in bold red, and the second-best results are highlighted in bold blue.}
    \label{table1}
\end{table*}

\begin{table}[ht]
    \centering
    \footnotesize
    \renewcommand{\arraystretch}{1}
    \setlength{\tabcolsep}{3pt}
    \scalebox{1.06}{
    \begin{tabular}{lcccc}
        \toprule
        \multirow{2}{*}{Methods} & \multicolumn{2}{c}{LSRW-Huawei~\cite{Hai_2023} } & \multicolumn{2}{c}{LSRW-Nikon~\cite{Hai_2023}} \\
        \cmidrule(lr){2-3} \cmidrule(lr){4-5}
        & PSNR & SSIM & PSNR & SSIM \\
        \midrule
        SID~\cite{Chen2019SeeingMI} & 17.47 & 0.652 & 16.33 & 0.613 \\
        RF~\cite{Kosugi_Yamasaki_2020} & 19.05 & 0.637 & 18.77 & 0.630 \\
        UFormer~\cite{wang2021uformergeneralushapedtransformer} & 19.77 & 0.643 & 19.77 & 0.643 \\
        EnGAN~\cite{Jiang2019EnlightenGANDL} & 20.22 & 0.701 & 20.71 & 0.659 \\ 
        Restormer~\cite{204} & \textcolor{blue}{22.61} & \textcolor{blue}{0.725} & \textcolor{blue}{21.20} &\textcolor{blue}{0.677} \\
        Retinexformer~\cite{205} & 19.57 & 0.578 & - & - \\
        LIME~\cite{7782813} & 17.00 & 0.382 & 13.53 & 0.332 \\
        MF~\cite{fu2016fusion} & 18.26 & 0.428 & 15.44 & 0.400 \\
        NPE~\cite{wang2013naturalness} & 17.08 & 0.391 & 14.86 & 0.374 \\
        SRIE~\cite{fu2016weighted} & 13.42 & 0.428 & 13.26 & 0.140 \\
        RetinexNet~\cite{wei2018deep} & 19.98 & 0.688 & 19.86 & 0.650 \\
        Kind~\cite{ref34} & 16.58 & 0.569 & 11.52 & 0.383 \\
        Kind++~\cite{zhang2021beyond} & 15.43 & 0.570 & 14.79 & 0.475 \\
        MIRNet~\cite{ref1839} & 19.98 & 0.609 & 17.10 & 0.502 \\
        SNR-Net~\cite{inproceedings} & 20.67 & 0.591 & 17.54 & 0.482 \\
        Bread~\cite{hu2021lowlightimageenhancementbreaking} & 19.20 & 0.618 & 14.70 & 0.487 \\
        LightenDiffusion~\cite{jiang2024lightendiffusionunsupervisedlowlightimage} & 18.56 & 0.539 & - & - \\
        3DLUT\cite{zeng2020lut}  & 18.12 & 0.659 & 17.81 & 0.629 \\
        Sparse\cite{9328179} & 20.33 & 0.699 & 20.19 & 0.657\\
        RUAS\cite{liu2021ruas}   & 20.46 & 0.704 & 20.88 & 0.664\\
        DRBN~\cite{Yang_2020_CVPR}   & 20.61 & 0.710 & 21.07 & 0.670\\
        \toprule
        \textbf{Ours} & \textcolor{red}{\textbf{23.55}} & \textcolor{red}{\textbf{0.741}} & \textcolor{red}{\textbf{22.62}} & \textcolor{red}{\textbf{0.700}} \\
        \bottomrule
    \end{tabular}%
    }
    
 \vspace{-0.8em}
    
\caption{\textbf{Quantitative comparison of the LSRW-Huawei and LSRW-Nikon datasets.} The best results are highlighted in bold red, and the second-best results are highlighted in bold blue.}
\label{table2}
\end{table}

\begin{figure*}[ht]
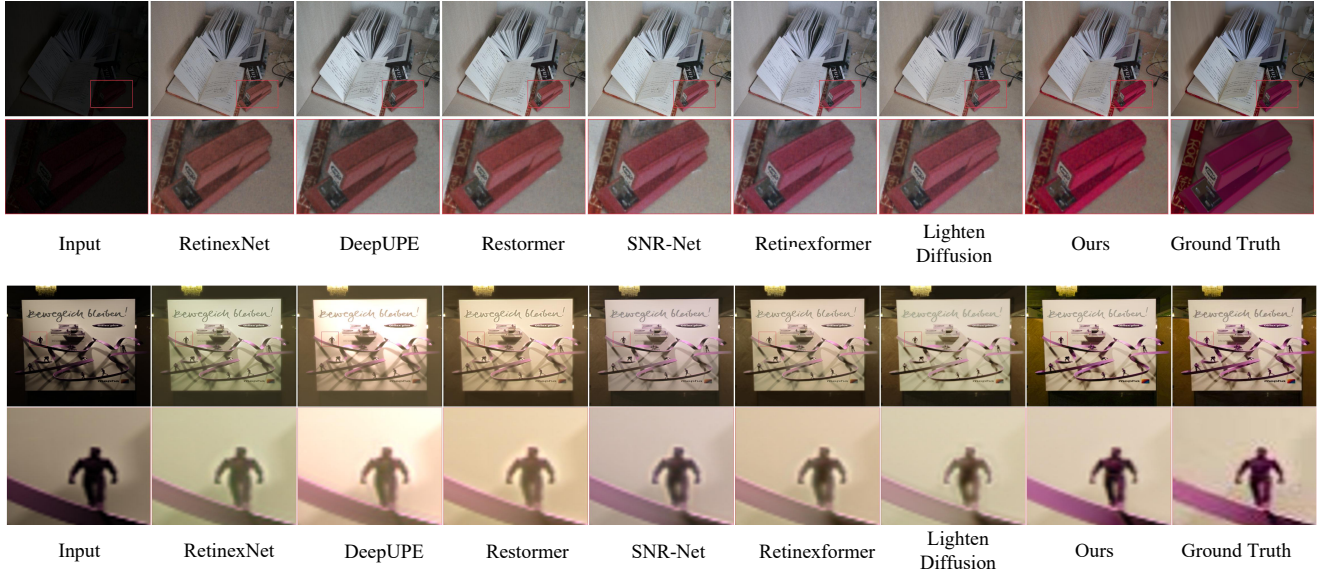

\vspace{-1em}
  \centering
  \includegraphics[width=\textwidth]{duibitu777_embed.pdf}
  \includegraphics[width=\textwidth]{duibitu11.pdf}
  \vspace{-1.12em}
  \caption{\textbf{Qualitative comparison on LOLv1 (first row) and LOLv2-Synthetic(second row) .}  It can be seen that the proposed method significantly improves image clarity, and the colours are closer to reality.}
  \label{fig:yourlabel2}
  \vspace{-1.4em}
\end{figure*}

\begin{figure}[ht]
\begin{center}
\hspace{-1em}
\scalebox{0.59}{
\begin{tabular}{c@{ } c@{ } c@{ }}
    \includegraphics[trim={500 120 310 510},clip,width=.26\textwidth]{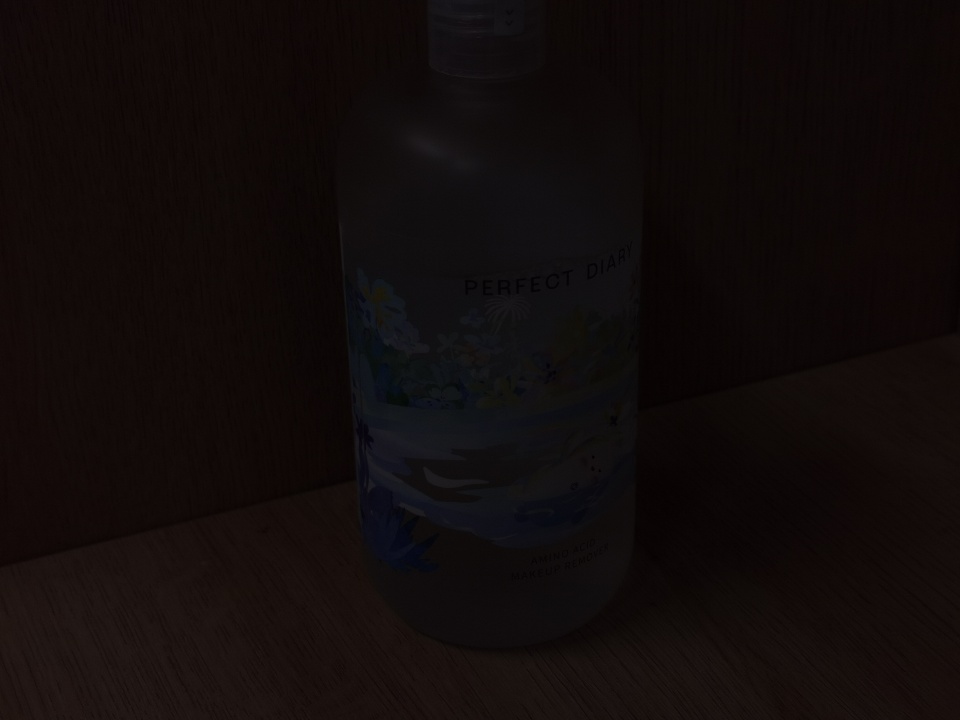} & 
    \includegraphics[trim={500 120 310 510},clip,width=.26\textwidth]{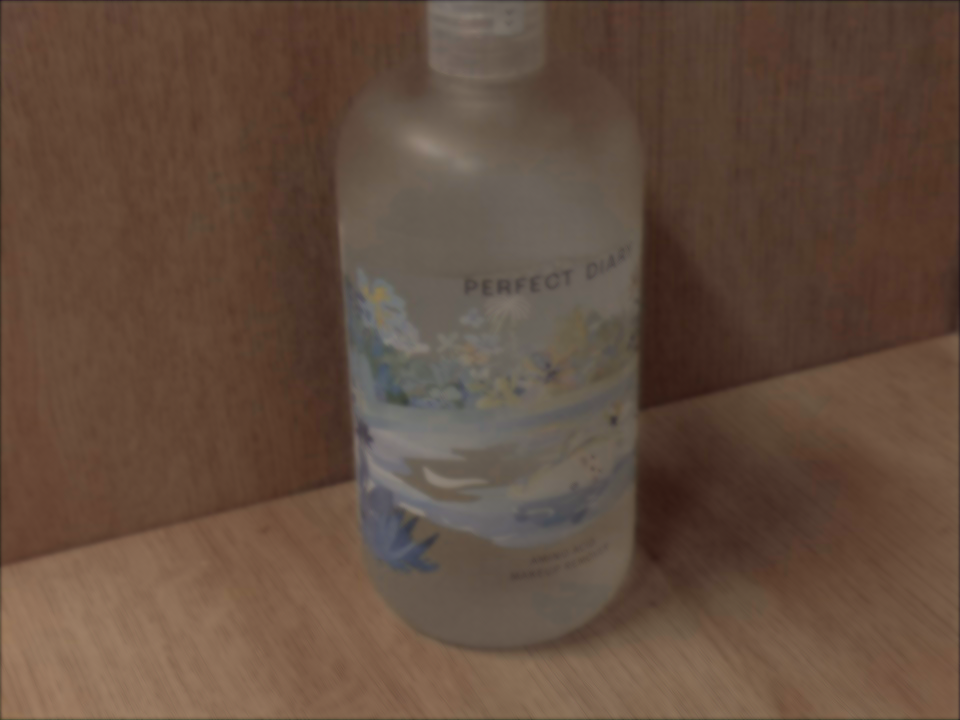} & 
    \includegraphics[trim={500 120 310 510},clip,width=.26\textwidth]{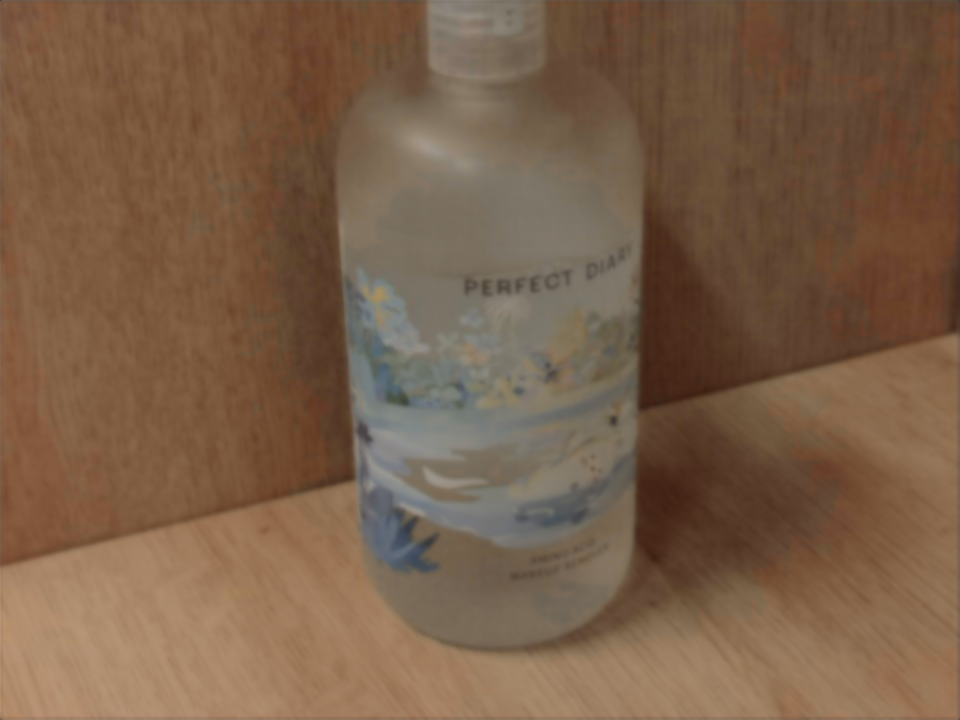} \\

    \large{PSNR} & \large{19.98 dB} & \large{22.61 dB} \\
    \large{Input Image}  & \large{RetinexNet~\cite{wei2018deep}} & \large{Restormer~\cite{204}} \\

    \includegraphics[trim={500 120 310 510},clip,width=.26\textwidth]{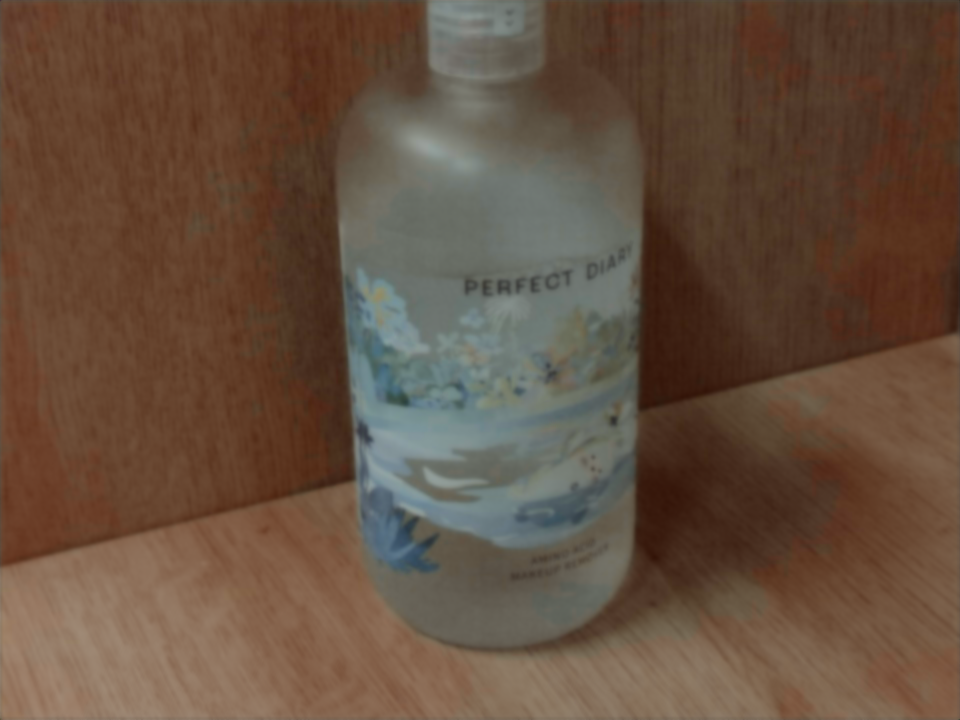} & 
    \includegraphics[trim={500 120 310 510},clip,width=.26\textwidth]{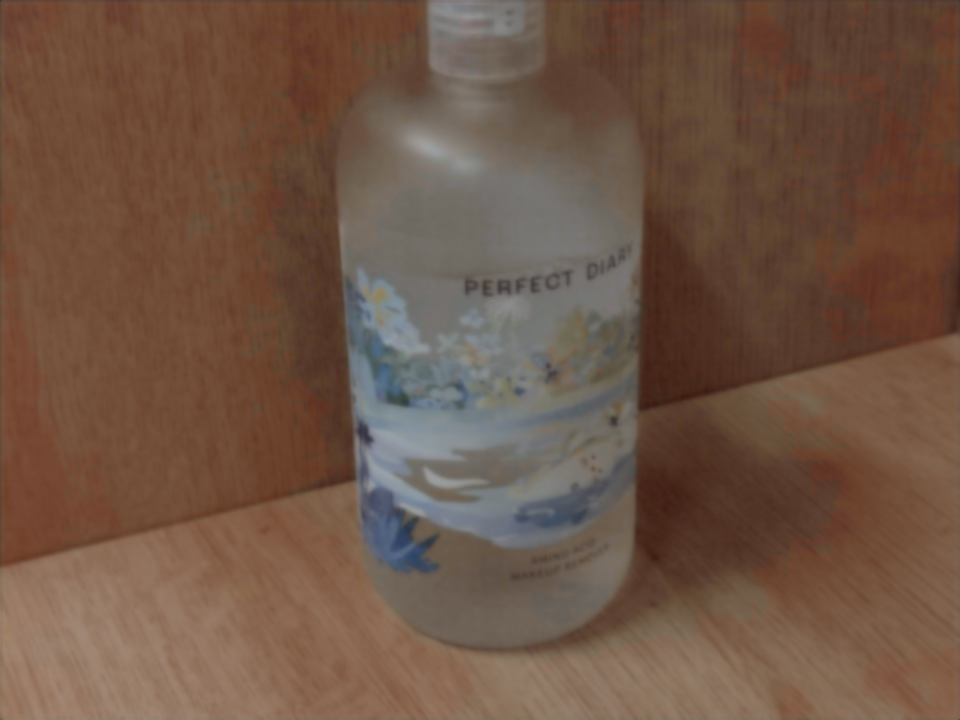} &  
    \includegraphics[trim={500 120 310 510},clip,width=.26\textwidth]{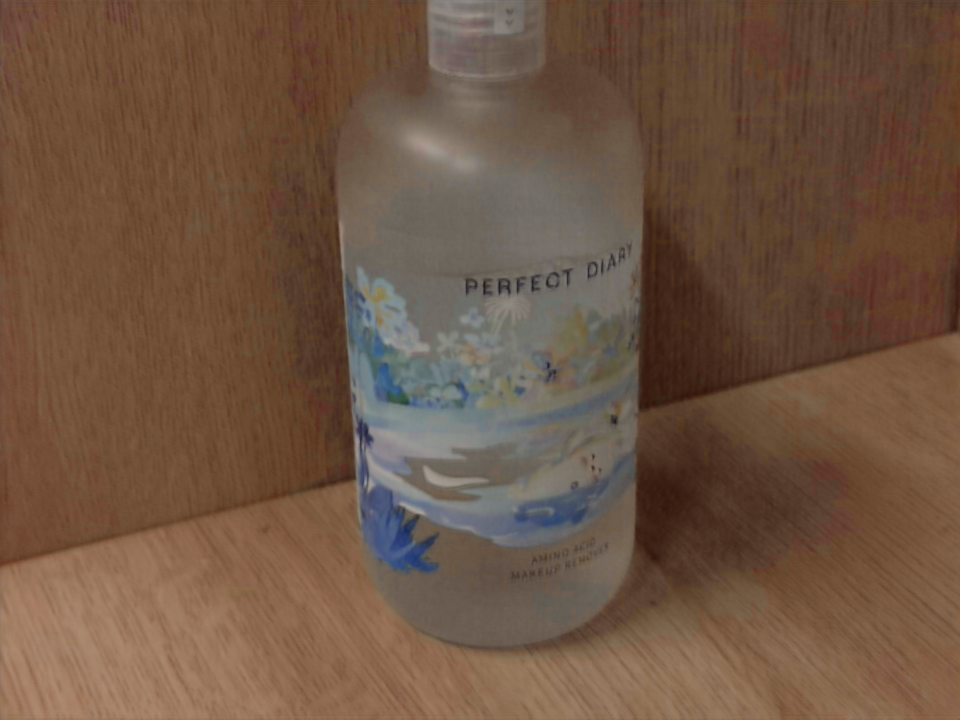} \\

    \large{20.67 dB} & \large{19.57 dB} & \large{23.55 dB} \\
    \large{SNR-Net~\cite{inproceedings}}  & \large{Retinexformer~\cite{205}} & \large{\textbf{Ours}}
\end{tabular}

}
\end{center}
\vspace{-1.4em}
\caption{\textbf{Image enhancement example.} Qualitative comparison on LSRW-Huawei. Our network clearly restores the fine details of the mineral water text in the image.}
\label{fig:kqs}
\vspace{-2 em}
\end{figure}

To evaluate the effectiveness of GM-MoE system, five prominent LLIE datasets were employed: LOL-v1~\cite{wei2018deep}, LOLv2-Real~\cite{9328179}, LOLv2-Synthetic~\cite{9328179}, LSRW-Huawei~\cite{Hai_2023}, and LSRW-Nikon~\cite{Hai_2023}. Specifically, LOL-v1 contains 485 training pairs and 15 test pairs captured from real scenes under different exposure times. LOLv2-Real, which includes 689 training pairs and 100 test pairs collected by adjusting exposure time and ISO, and LOLv2-Synthetic, which is generated by analyzing the lighting distribution of low-light images and contains 900 training pairs and 100 test pairs. The LSRW-Huawei and LSRW-Nikon datasets each contain several real low-light images captured by devices in real-world scenes.



\noindent\textbf{Implementation Details}: The GM-MoE was developed using the PyTorch framework and trained on an NVIDIA 4090 GPU. The training process began with an initial learning rate of \(1.0 \times 10^{-3}\), which was managed using a multi-step scheduler. The Adam~\cite{Loshchilov2017DecoupledWD} optimizer, configured with a momentum parameter of 0.9, was used for optimization. During training, input images were resized to \(256 \times 256\) pixels and subjected to data augmentation techniques, including random rotations and flips, to enhance model generalization. A batch size of 4 was maintained, and the training regimen consisted of a total of \(2.0 \times 10^6\) iterations. Performance was evaluated using Peak Signal-to-Noise Ratio (PSNR) and Structural Similarity Index (SSIM)~\cite{1284395} as the primary metrics.

\subsection{Low-light Image Enhancement}
\noindent \textbf{Quantitative Results.} Among the deep learning methods for low-light image enhancement are  SID~\cite{Chen2019SeeingMI},
RF~\cite{Kosugi_Yamasaki_2020},
UFormer~\cite{wang2021uformergeneralushapedtransformer},
EnGAN~\cite{Jiang2019EnlightenGANDL}, Restormer~\cite{204}, Retinexformer~\cite{205}, DeepUPE~\cite{Wang_2019_CVPR}, RetinexNet~\cite{wei2018deep}, Kind~\cite{ref34}, Kind++~\cite{zhang2021beyond}, MIRNet~\cite{ref1839}, SNR-Net~\cite{inproceedings}, Bread~\cite{hu2021lowlightimageenhancementbreaking}, DPEC~\cite{wang2024dpecdualpatherrorcompensation}, as well as  traditional methods such as, LIME~\cite{7782813}, MF~\cite{fu2016fusion}, NPE~\cite{wang2013naturalness}, and SRIE~\cite{fu2016weighted}, GM-MoE achieved an overall performance improvement on all datasets. Tab.~\ref{table1} shows the quantitative comparison results of GM-MoE with a variety of SOTA image enhancement algorithms. GM-MoE achieved PSNR improvements of 0.9, 0.76, and 0.11 dB on the LOL-v1~\cite{wei2018deep}, LOLv2-Real~\cite{9328179}, and LOLv2-Synthetic~\cite{9328179} datasets, respectively, compared to the second-ranked model in each dataset, significantly improving image quality. This showed that GM-MoE performed well on these classic datasets, consistently outperforming other methods in terms of detail recovery and color enhancement, both in synthetic and real low-light conditions.
In addition, as shown in Tab.~\ref{table2}, GM-MoE also achieved significant improvements over other SOTA methods on the LSRW-Huawei~\cite{Hai_2023} and LSRW-Nikon~\cite{Hai_2023} datasets, achieving PSNR improvements of 0.94 dB and 1.42 dB over the next best method, Restormer~\cite{204}, respectively. These two datasets contained a large amount of noise and artifacts, and GM-MoE was able to effectively reduce artifacts and recover image details and features in high-noise environments. For more results please refer to supplementary material.


\begin{figure}[ht] 
\centering 
\includegraphics[width=\columnwidth]{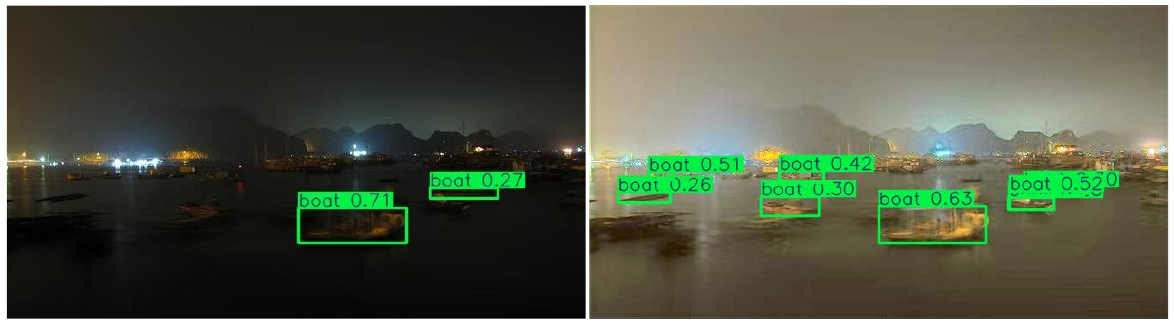} 
\caption{A visual comparison of the object detection task in low-light scenes (left) and scenes enhanced by GM-MoE (right).}.
\label{fig:mubiao} 
\vspace{-2.0em}
\end{figure}

\noindent\textbf{Qualitative Results.}  
 A visual comparison of the GM-MoE and the other algorithms is shown in Fig.~\ref{fig:yourlabel2}, and~\ref{fig:kqs} (Zoom in for better visualization.). Previous methods exhibited poor edge detail processing, with some blurring effects and noise, as shown in Fig.~\ref{fig:kqs} for RetinexNet~\cite{wei2018deep}, Retinexformer~\cite{205}, and SNR-Net~\cite{inproceedings}. Moreover, multiple networks had issues with color distortion, as shown in Fig.~\ref{fig:yourlabel2} for RetinexNet~\cite{wei2018deep} and Fig.~\ref{fig:kqs} for SNR-Net~\cite{inproceedings}. In addition, there were cases of underexposure or overexposure, as seen in DeepUPE~\cite{Wang_2019_CVPR} in Fig.~\ref{fig:yourlabel2}. In contrast, our work effectively restored colors, efficiently restored details, extracted shallow features, significantly reduced noise, and reliably preserved colors. As can be seen, our method outperformed other supervised and unsupervised algorithms across various scenarios and excelled in multiple metrics. The left and right parts of Fig.~\ref{fig:mubiao} show the performance of object detection in low-light scenes (left) and photos enhanced by GM-MoE, respectively.

\begin{figure*}[ht]
  \centering
  \includegraphics[width=\textwidth]{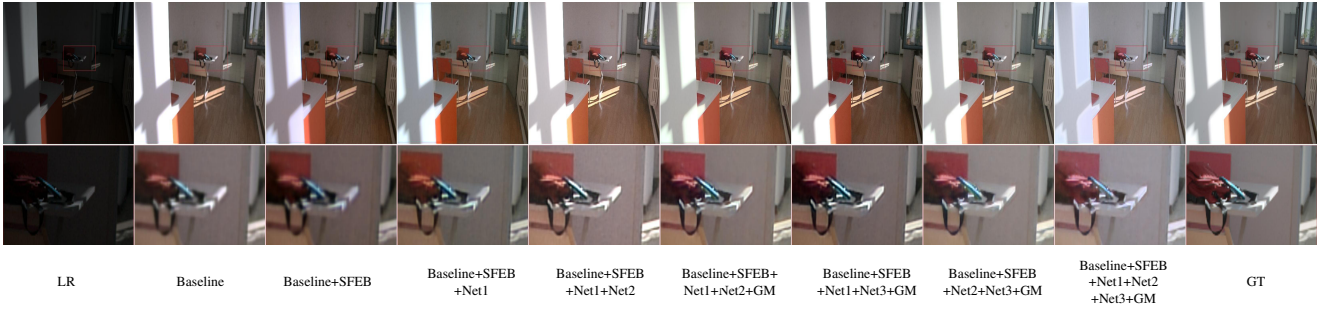}
  \caption{\textbf{Qualitative comparison of different ablation settings.} It can be observed that the final configuration (Baseline+SFEB+Net1+Net2+Net3+GM) produces the best results.}
  \label{fig:xiaorong}
  \vspace{-1.4em}
\end{figure*}

\subsection{Ablation Study.}
We incrementally added modules to the baseline model to assess their contributions and to conduct ablation studies on the LOL-v2-real~\cite{9328179} and LOL-v2-syn~\cite{9328179} datasets. The results are reported in Tab.~\ref{table3}.

\noindent\textbf{The Effectiveness of the Shallow Feature Extraction Network.} To verify the effectiveness of the shallow feature extraction network, we first introduced SFEB into the baseline model. Comparing the results of Experiment 1 (baseline model) with those of Experiment 2 (baseline + SFEB), we found that, on the LOL-v2-syn dataset, PSNR improved by \textbf{3.09 dB} and SSIM improved by 0.0215. This shows that SFEB can effectively extract the shallow features of an image, providing better feature input for the subsequent GM-MoE module.

\noindent\textbf{Is the Effectiveness Among Multiple Experts and the Performance Complementary?} To explore the role of each expert module (Expert1,Expert2, Expert3) and its complementary role in image enhancement, we gradually added each subnetwork to the model and performed ablation experiments. After adding \textbf{Expert1}, compared with Experiment 2, PSNR increased by \textbf{1.08 dB} (LOL-v2-real) .
After adding \textbf{Expert2}, compared with Experiment 3, PSNR increased by 0.76 dB (LOL-v2-real) and 0.79 dB (LOL-v2-syn), and SSIM increased by 0.0575 and 0.0891, respectively. Subsequently, we removed Expert1, Expert2, and Expert3 respectively through ablation experiments (Experiments 5--7), and a decrease in performance was observed in all cases. This shows that the expert modules can work together in synergy after being integrated into a module to solve the image enhancement problems they are designed for, improving the overall image restoration effect.

\noindent\textbf{Does the Gated Weight Generation Network Improve Generalization Ability?} In the complete model, we further introduce a gated weight generation network. Compared to models without this mechanism, it dynamically adjusts the weights of individual experts based on images from different data domains, thereby enhancing cross-domain generalization capability.

\begin{table}[h]
    \scriptsize
    \begin{center}
    \resizebox{\columnwidth}{!}{ 
    \setlength{\tabcolsep}{2pt} 
    \scalebox{1.1}{
    \begin{tabular}{lcccccc@{\hspace{10pt}}cc@{\hspace{10pt}}cc} 
        \toprule
        \multirow{2}{*}{ID} & \multirow{2}{*}{Baseline} & \multirow{2}{*}{SFEB} & \multirow{2}{*}{Net1} & \multirow{2}{*}{Net2} & \multirow{2}{*}{Net3} & \multirow{2}{*}{GM} & \multicolumn{2}{c}{LOL-v2-real~\cite{9328179}} & \multicolumn{2}{c}{LOL-v2-syn~\cite{9328179}} \\
        \cmidrule(lr){8-9} \cmidrule(lr){10-11}
        & & & & & & & \scalebox{1.1}{PSNR} & \scalebox{1.1}{SSIM} & \scalebox{1.1}{PSNR} & \scalebox{1.1}{SSIM} \\
        \midrule
        1 & \checkmark & & & & & & \scalebox{1.3}{19.45} & \scalebox{1.3}{0.7079} & \scalebox{1.3}{20.35} & \scalebox{1.3}{0.7431} \\
        2 & \checkmark & \checkmark & & & & & \scalebox{1.3}{20.27} & \scalebox{1.3}{0.7236} & \scalebox{1.3}{23.44} & \scalebox{1.3}{0.7646} \\
        3 & \checkmark & \checkmark & \checkmark & & & & \scalebox{1.3}{21.35} & \scalebox{1.3}{0.7446} & \scalebox{1.3}{24.35} & \scalebox{1.3}{0.8436} \\
        4 & \checkmark & \checkmark & \checkmark & \checkmark & & & \scalebox{1.3}{22.11} & \scalebox{1.3}{0.8021} & \scalebox{1.3}{25.14} & \scalebox{1.3}{0.9327} \\
        5 & \checkmark & \checkmark & \checkmark & \checkmark & & \checkmark & \scalebox{1.3}{23.23} & \scalebox{1.3}{0.8045} & \scalebox{1.3}{26.08} & \scalebox{1.3}{0.9351} \\
        6 & \checkmark & \checkmark & \checkmark & & \checkmark & \checkmark & \scalebox{1.3}{23.31} & \scalebox{1.3}{0.8054} & \scalebox{1.3}{26.12} & \scalebox{1.3}{0.9362} \\
        7 & \checkmark & \checkmark & & \checkmark & \checkmark & \checkmark & \scalebox{1.3}{23.35} & \scalebox{1.3}{0.8055} & \scalebox{1.3}{26.15} & \scalebox{1.3}{0.9366} \\
        8 & \checkmark & \checkmark & \checkmark & \checkmark & \checkmark & \checkmark & \textcolor{red}{\textbf{\scalebox{1.3}{23.65}}} & \textcolor{red}{\textbf{\scalebox{1.3}{0.8060}}} & \textcolor{red}{\textbf{\scalebox{1.3}{26.29}}} & \textcolor{red}{\textbf{\scalebox{1.3}{0.9371}}} \\
        \bottomrule
    \end{tabular}%
    }
    }
    \vspace{-2em}
    \end{center}
\caption{\textbf{Ablation study results on LOL-v2-real and LOL-v2-syn datasets.} The best results are highlighted in bold red.}
\label{table3}
\end{table}

\noindent\textbf{Qualitative results.}
As shown in the ablation study results in Fig.~\ref{fig:xiaorong}, images generated using only the baseline model exhibit noticeable blurring and loss of detail. The sequential integration of SFEB, Net1, Net2, Net3, and the gated weight generation module (GM) progressively enhances the model’s ability to recover low-light images. Each component contributes uniquely to the overall performance, and the complete model achieves the best results.

\section{Conclusion}
This paper proposes the GM-MoE framework, which dynamically balances three expert subnetworks based on input image features, simultaneously addressing color bias, detail loss, and insufficient illumination in low-light images while ensuring strong generalization across diverse data domains. Extensive quantitative and qualitative evaluations on five benchmark datasets demonstrate that GM-MoE outperforms existing methods in both PSNR and SSIM metrics. Future work will focus on real-time enhancement and adaptive optimization.
\newpage
\clearpage
\setcounter{page}{1}
\maketitlesupplementary

\appendix
\section{Appendix Section}
\subsection{Visualization of the results of a comparison experiment}

\begin{figure*}[ht]
    \centering

    \begin{subfigure}{0.23\linewidth}
        \centering
        \includegraphics[width=\linewidth]{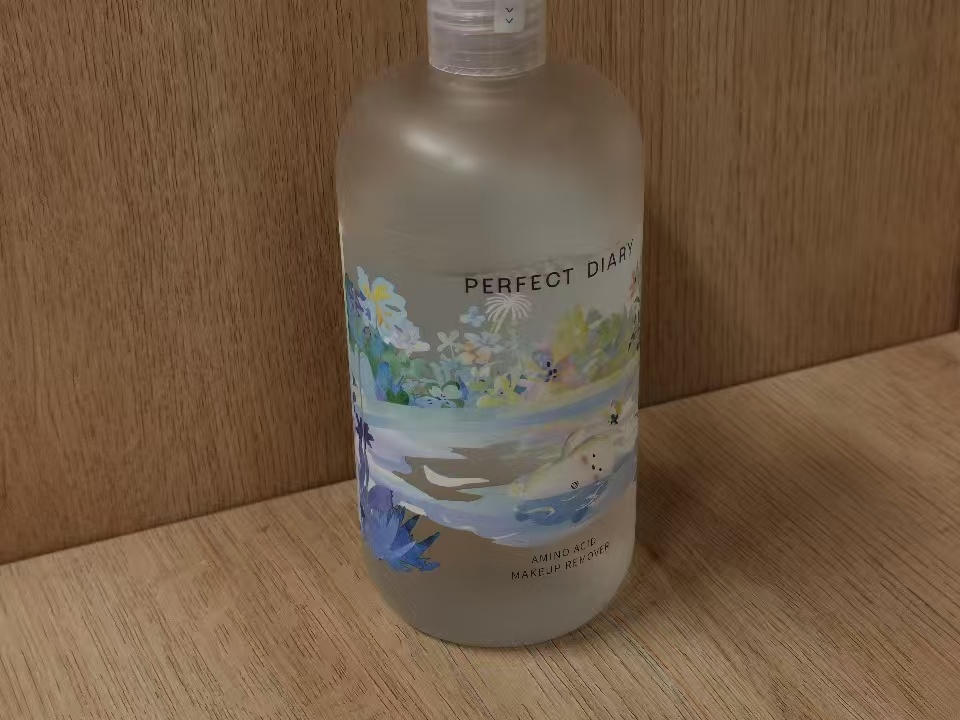}
        \caption{Ground Truth}
        \label{fig:gt}
    \end{subfigure}
    \hfill
    \begin{subfigure}{0.23\linewidth}
        \centering
        \includegraphics[width=\linewidth]{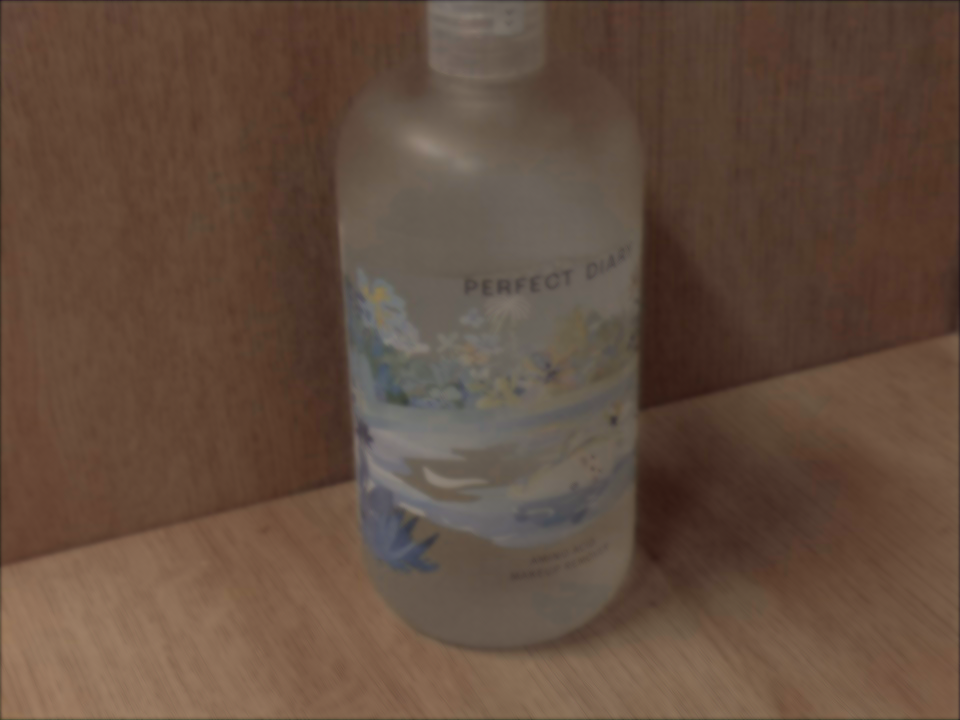}
        \caption{RetinexNet~\cite{wei2018deep}}
        \label{fig:retinexnet}
    \end{subfigure}
    \hfill
    \begin{subfigure}{0.23\linewidth}
        \centering
        \includegraphics[width=\linewidth]{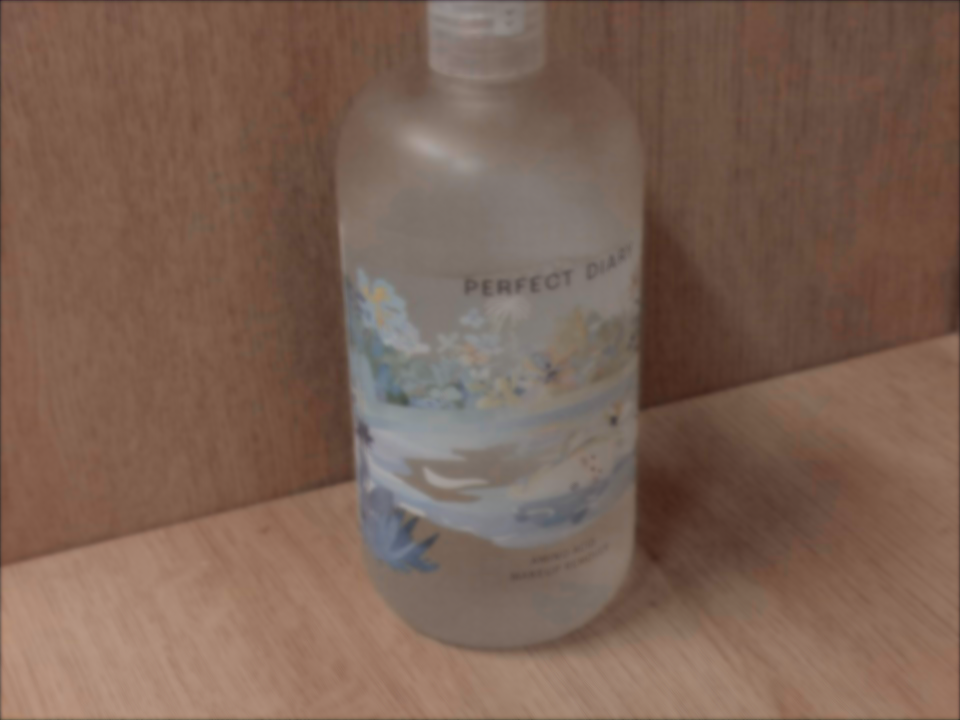}
        \caption{DeepUPE~\cite{Wang_2019_CVPR}}
        \label{fig:deepupe}
    \end{subfigure}
    \hfill
    \begin{subfigure}{0.23\linewidth}
        \centering
        \includegraphics[width=\linewidth]{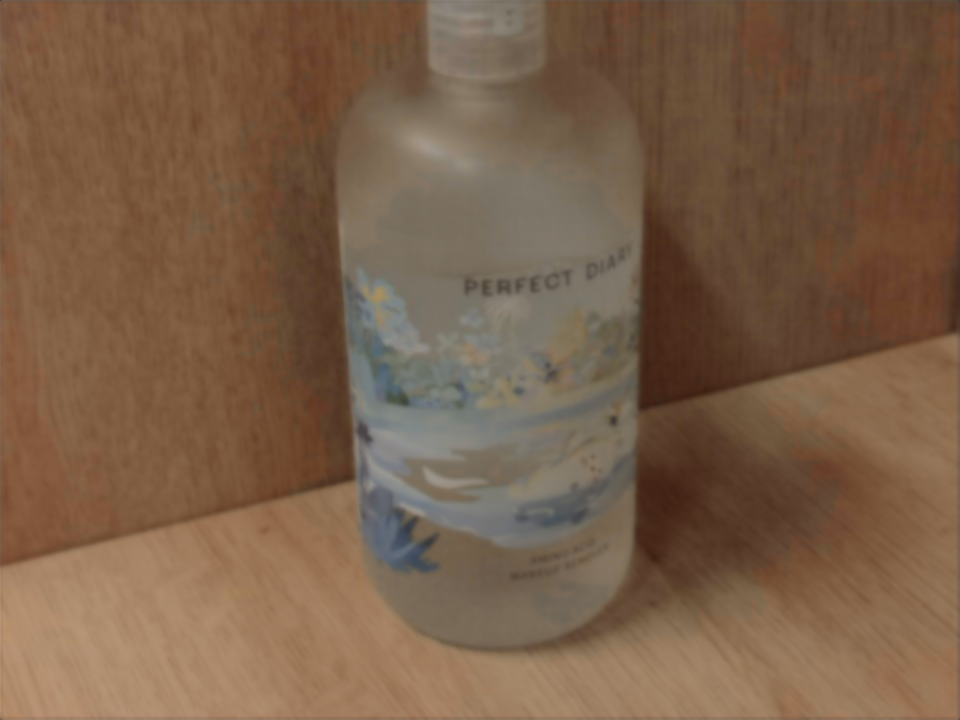}
        \caption{Restormer~\cite{204}}
        \label{fig:restormer}
    \end{subfigure}

    \medskip

    \begin{subfigure}{0.23\linewidth}
        \centering
        \includegraphics[width=\linewidth]{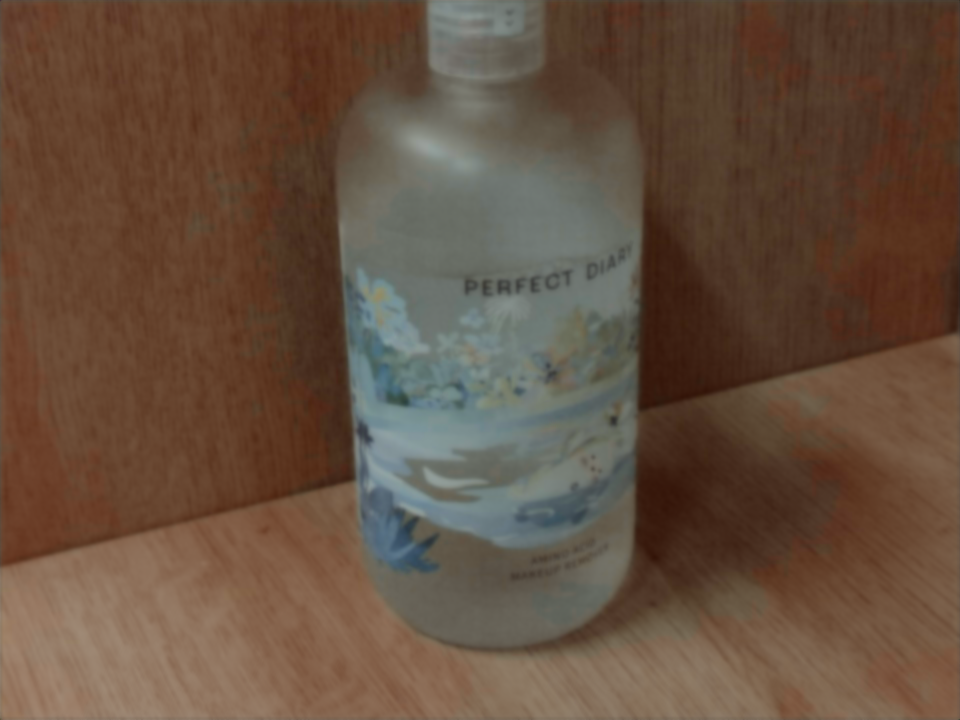}
        \caption{SNR-Net~\cite{inproceedings}}
        \label{fig:snrnet}
    \end{subfigure}
    \hfill
    \begin{subfigure}{0.23\linewidth}
        \centering
        \includegraphics[width=\linewidth]{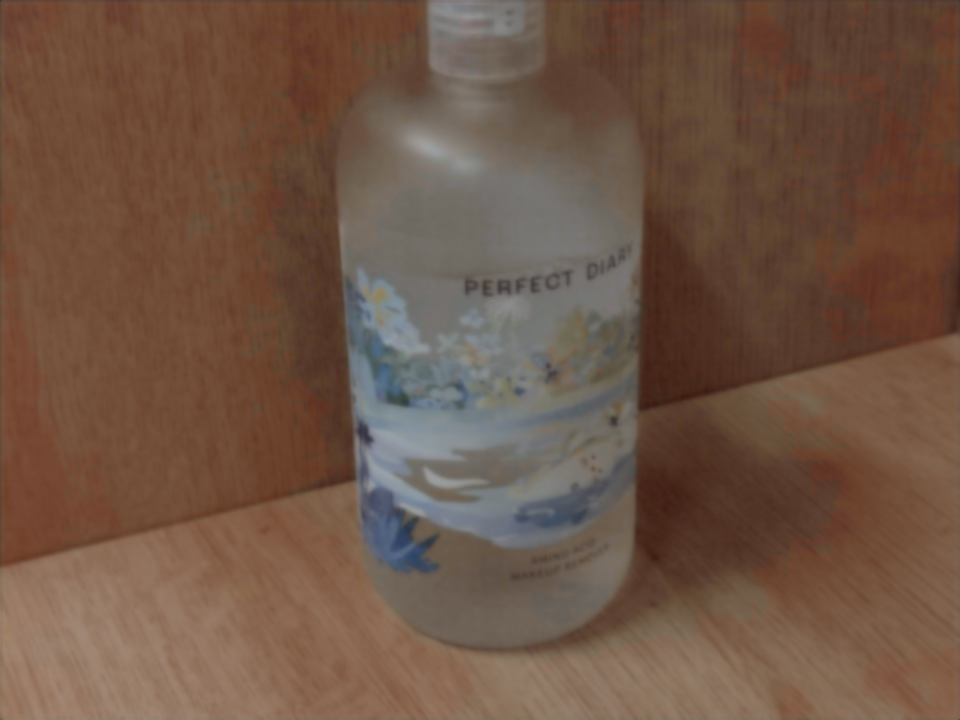}
        \caption{Retinexformer~\cite{205}}
        \label{fig:retinexformer}
    \end{subfigure}
    \hfill
    \begin{subfigure}{0.23\linewidth}
        \centering
        \includegraphics[width=\linewidth]{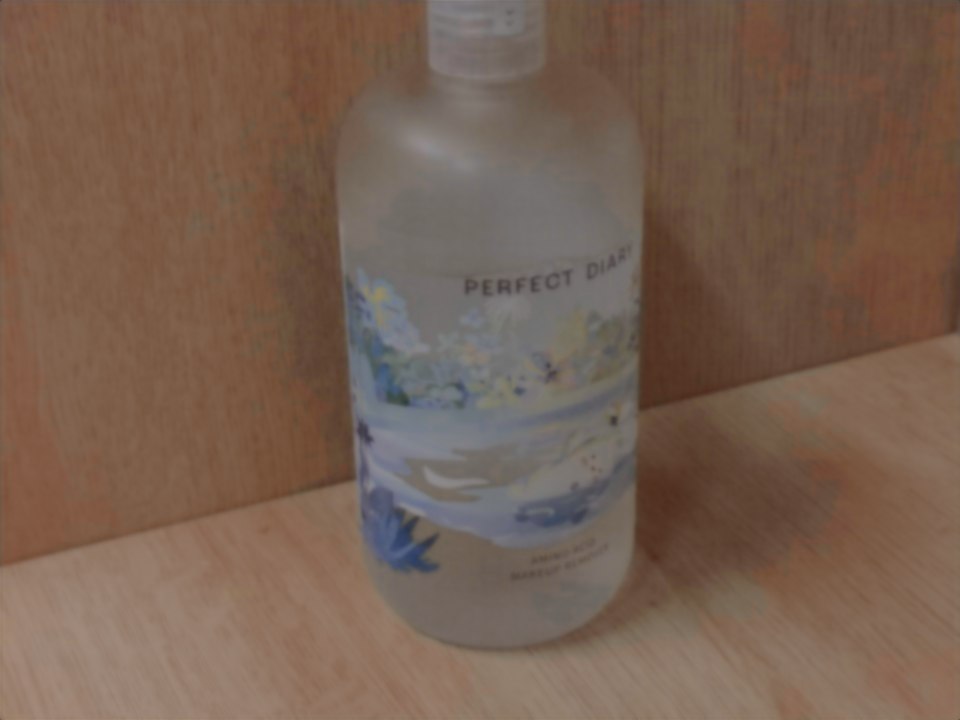}
        \caption{LightenDiffusion~\cite{jiang2024lightendiffusionunsupervisedlowlightimage}}
        \label{fig:ld1}
    \end{subfigure}
    \hfill
    \begin{subfigure}{0.23\linewidth}
        \centering
        \includegraphics[width=\linewidth]{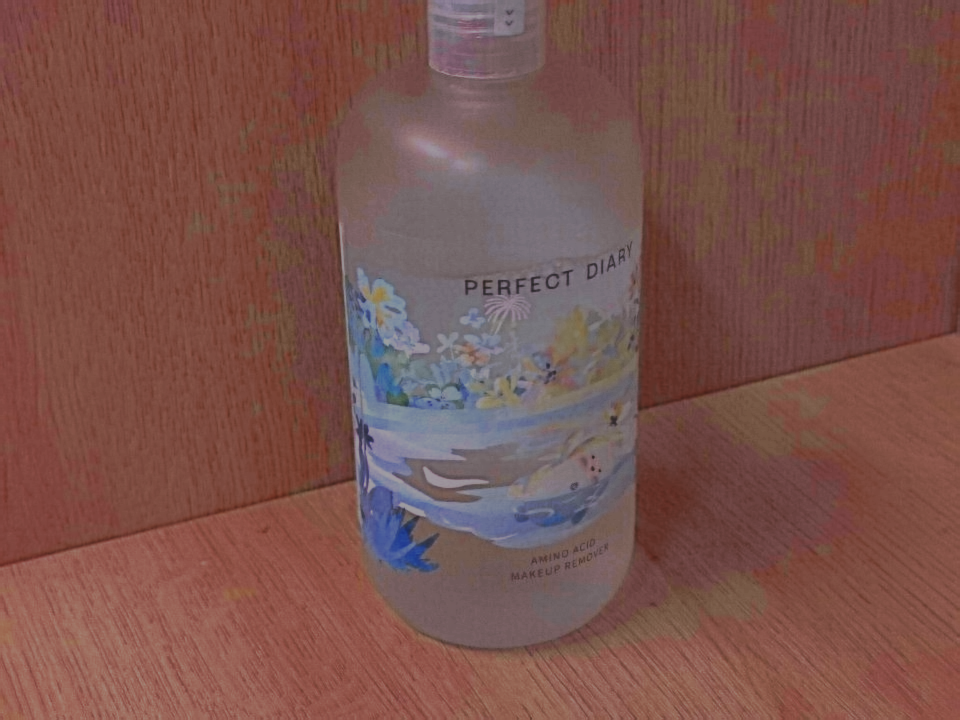}
        \caption{PairLIE~\cite{fu2023learning}}
        \label{fig:ld2}
    \end{subfigure}

    \medskip
    \begin{subfigure}{0.23\linewidth}
        \centering
        \includegraphics[width=\linewidth]{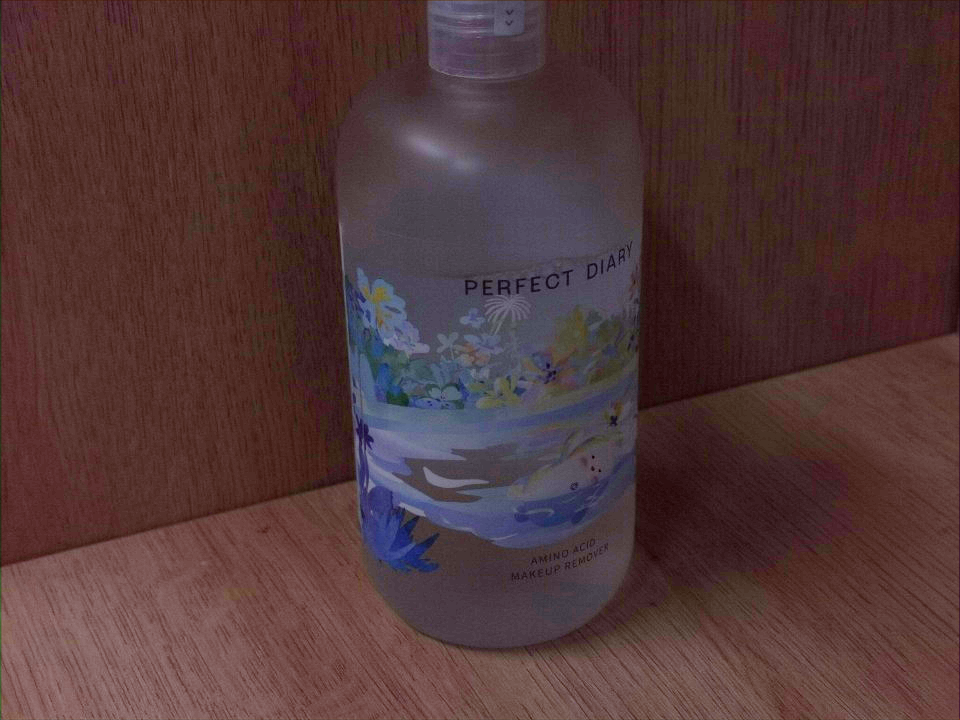}
        \caption{SCI~\cite{Ma_2022_CVPR}}
        \label{fig:ld3}
    \end{subfigure}
    \hfill
    \begin{subfigure}{0.23\linewidth}
        \centering
        \includegraphics[width=\linewidth]{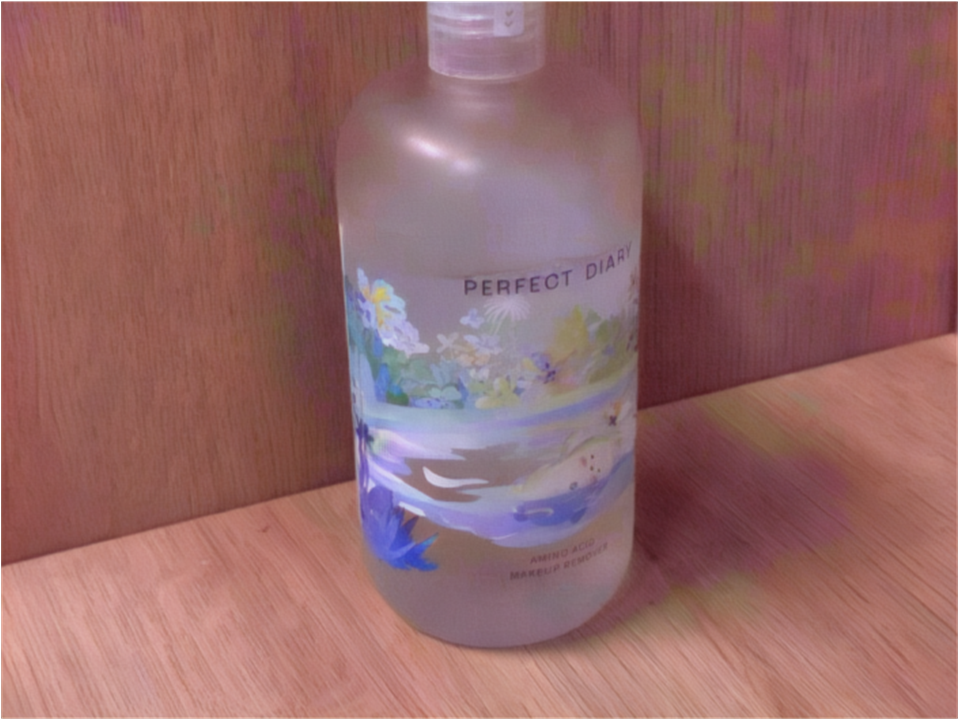}
        \caption{QuadPrior~\cite{Wang_2024_CVPR}}
        \label{fig:ld4}
    \end{subfigure}
    \hfill
    \begin{subfigure}{0.23\linewidth}
        \centering
        \includegraphics[width=\linewidth]{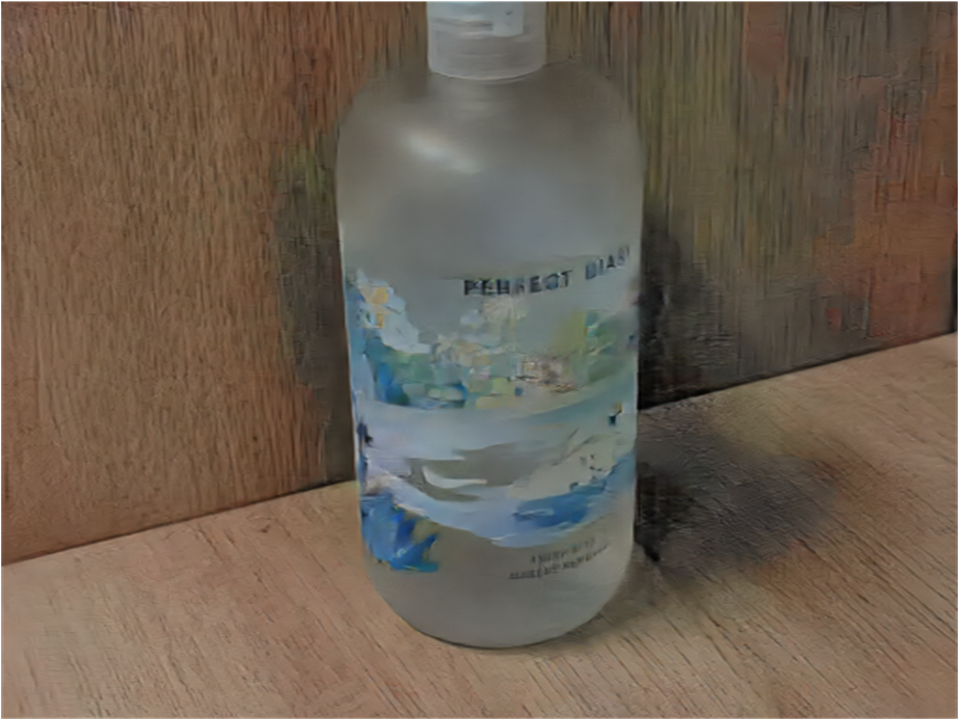}
        \caption{NeRCo~\cite{Yang_2023_ICCV}}
        \label{fig:ld5}
    \end{subfigure}
    \hfill
    \begin{subfigure}{0.23\linewidth}
        \centering
        \includegraphics[width=\linewidth]{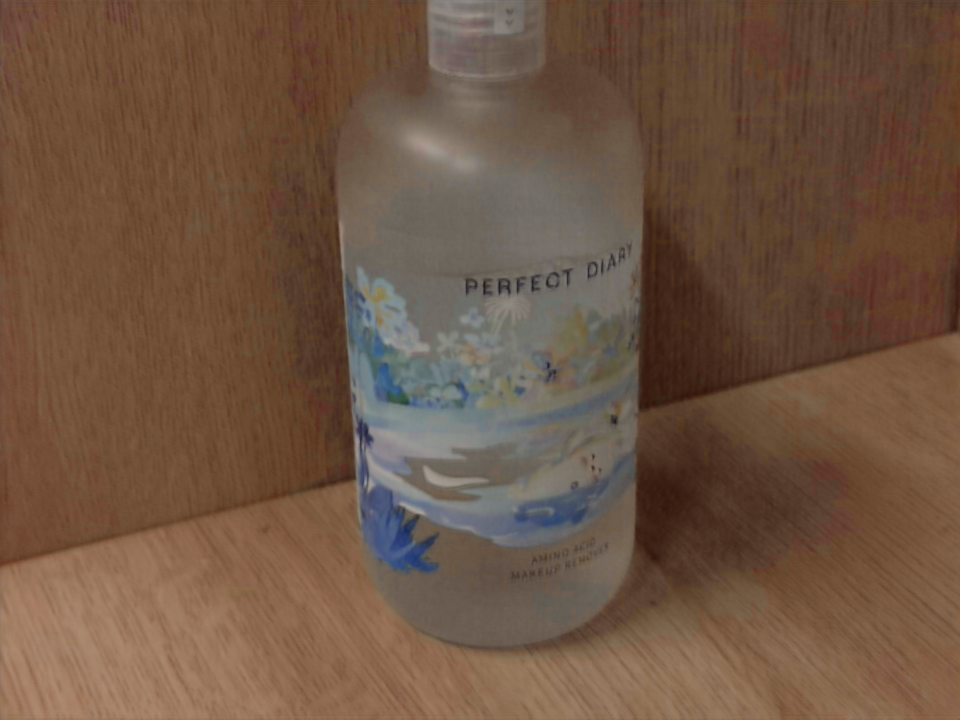}
        \caption{Ours (GM-MOE)}
        \label{fig:ours}
    \end{subfigure}

    \caption{\textbf{Visual comparison on the LSRW-Huawei dataset~\cite{Hai_2023}.} The models compared include RetinexNet, DeepUPE, Restormer, SNR-Net, Retinexformer, LightenDiffusion, PairLIE, SCI, QuadPrior, NeRCo, Ours (our model), and Ground Truth. Among them, GM-MOE achieves better enhancement compared to other models. Zoom in to see more details of the differences between models.}
    \label{fig:1}
\end{figure*}

\begin{figure}[h]
    \centering
    \begin{subfigure}{0.48\linewidth}
        \centering
        \includegraphics[width=\linewidth]{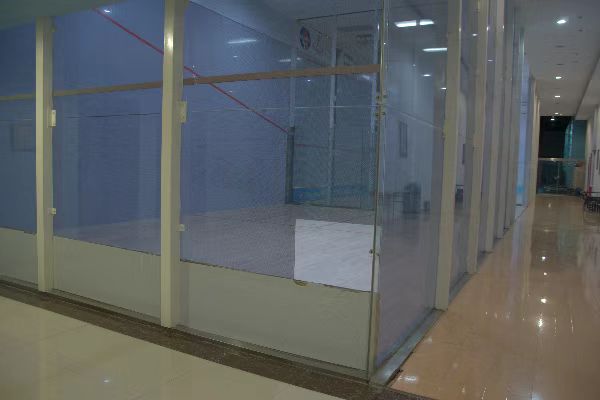}
        \captionsetup{font={footnotesize,bf,stretch=1},justification=raggedright}
        \caption{Ground Truth}
        \label{fig:huawei_input}
    \end{subfigure}
    \hfill
    \begin{subfigure}{0.48\linewidth}
        \centering
        \includegraphics[width=\linewidth]{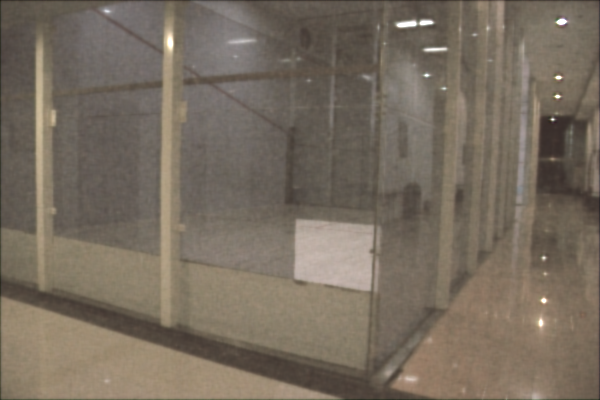}
        \captionsetup{font={footnotesize,bf,stretch=1},justification=raggedright}
        \caption{RetinexNet~\cite{wei2018deep}}
        \label{fig:huawei_deepupe}
    \end{subfigure}
    
    \medskip  
    
    \begin{subfigure}{0.48\linewidth}
        \centering
        \includegraphics[width=\linewidth]{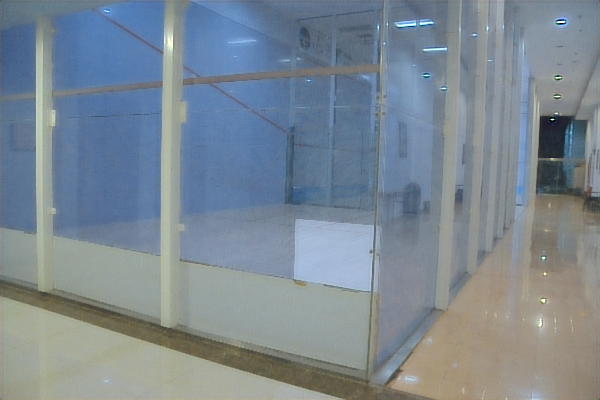}
        \captionsetup{font={footnotesize,bf,stretch=1},justification=raggedright}
        \caption{DeepUPE~\cite{Wang_2019_CVPR}}
        \label{fig:huawei_ours}
    \end{subfigure}
    \hfill
    \begin{subfigure}{0.48\linewidth}
        \centering
        \includegraphics[width=\linewidth]{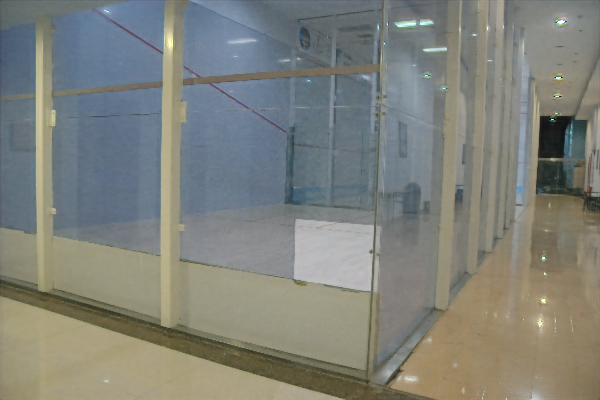}
        \captionsetup{font={footnotesize,bf,stretch=1},justification=raggedright}
        \caption{Restormer~\cite{204}}
        \label{fig:huawei_gt}
    \end{subfigure}

    \begin{subfigure}{0.48\linewidth}
        \centering
        \includegraphics[width=\linewidth]{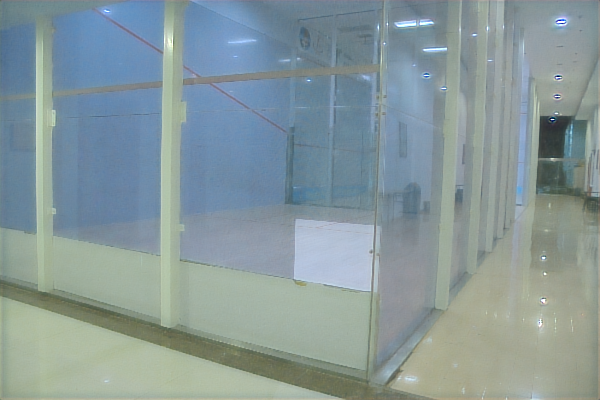}
        \captionsetup{font={footnotesize,bf,stretch=1},justification=raggedright}
        \caption{SNR-Net~\cite{inproceedings}}
        \label{fig:huawei_input}
    \end{subfigure}
    \hfill
    \begin{subfigure}{0.48\linewidth}
        \centering
        \includegraphics[width=\linewidth]{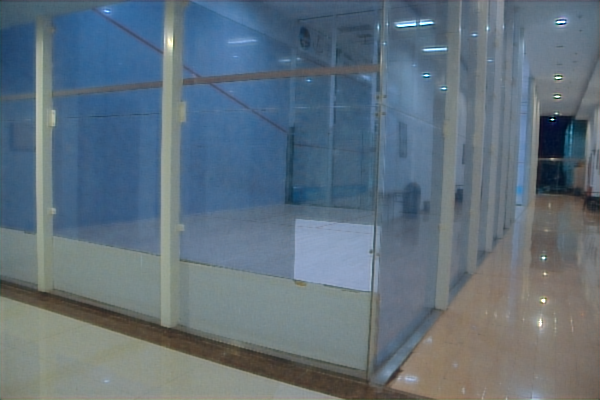}
        \captionsetup{font={footnotesize,bf,stretch=1},justification=raggedright}
        \caption{Retinexformer~\cite{205} }
        \label{fig:huawei_deepupe}
    \end{subfigure}
    
    \medskip  
    
    \begin{subfigure}{0.48\linewidth}
        \centering
        \includegraphics[width=\linewidth]{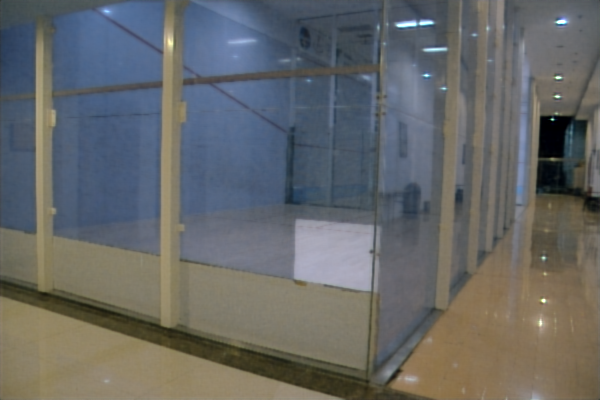}
        \captionsetup{font={footnotesize,bf,stretch=1},justification=raggedright}  \caption{LightenDiffusion~\cite{jiang2024lightendiffusionunsupervisedlowlightimage}}
        \label{fig:huawei_ours}
    \end{subfigure}
    \hfill
    \begin{subfigure}{0.48\linewidth}
        \centering
        \includegraphics[width=\linewidth]{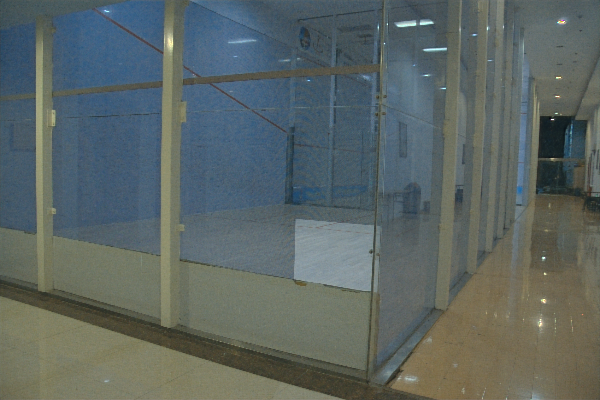}
        \captionsetup{font={footnotesize,bf,stretch=1},justification=raggedright}
        \caption{Ours(GM-MOE)}
        \label{fig:huawei_gt}
    \end{subfigure}

    
    \caption{\textbf{Visual comparison on the LOL-v2-Real
    dataset.}The enhanced effect of GM-MOE is better than other models in terms of detail processing.}
    \label{fig:lol}
\end{figure}

\begin{figure}[h]
    \centering
    \begin{subfigure}{0.48\linewidth}
        \centering
        \includegraphics[width=\linewidth]{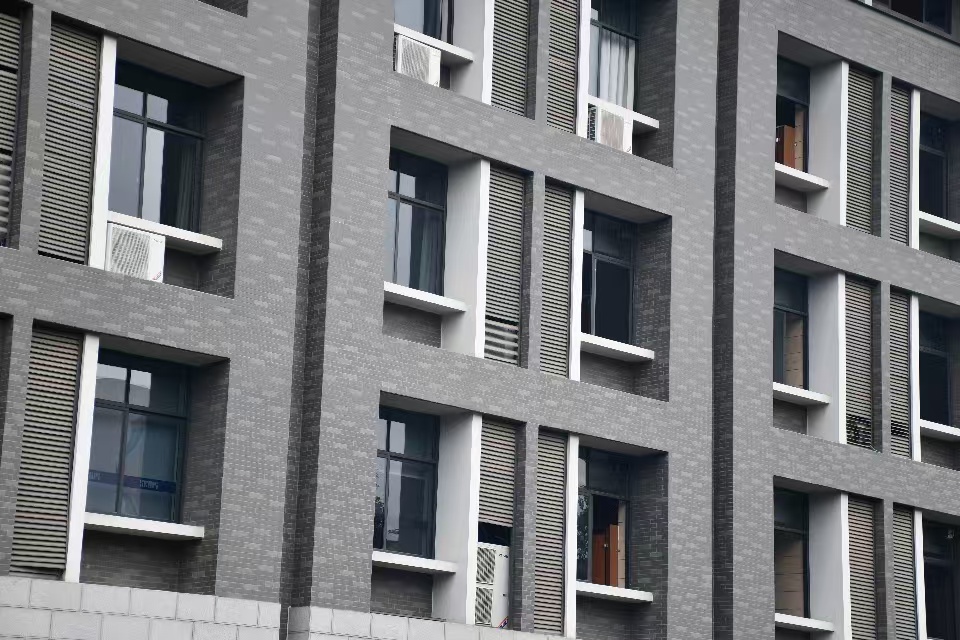}
        \captionsetup{font={footnotesize,bf,stretch=1},justification=raggedright}
        \caption{Ground Truth}
        \label{fig:huawei_input}
    \end{subfigure}
    \hfill
    \begin{subfigure}{0.48\linewidth}
        \centering
        \includegraphics[width=\linewidth]{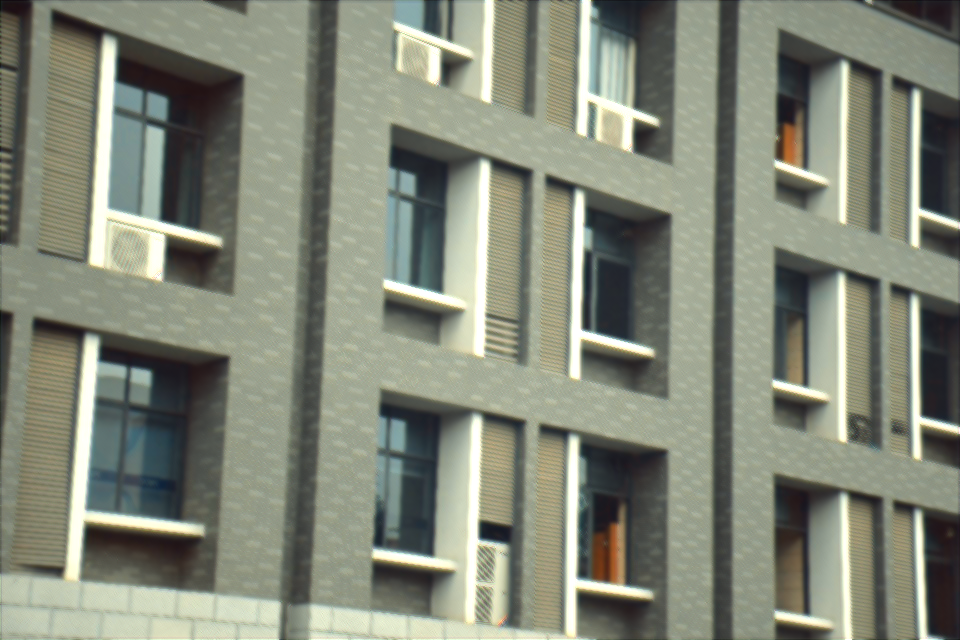}
        \captionsetup{font={footnotesize,bf,stretch=1},justification=raggedright}
        \caption{RetinexNet~\cite{wei2018deep}}
        \label{fig:huawei_deepupe}
    \end{subfigure}
    
    \medskip  
    
    \begin{subfigure}{0.48\linewidth}
        \centering
        \includegraphics[width=\linewidth]{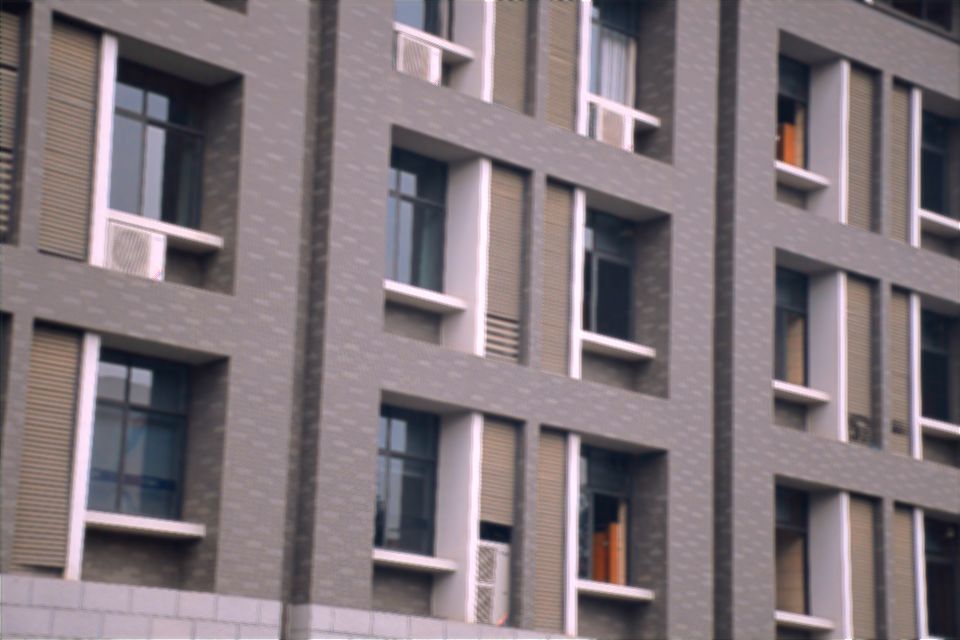}
        \captionsetup{font={footnotesize,bf,stretch=1},justification=raggedright}
        \caption{DeepUPE~\cite{Wang_2019_CVPR}}
        \label{fig:huawei_ours}
    \end{subfigure}
    \hfill
    \begin{subfigure}{0.48\linewidth}
        \centering
        \includegraphics[width=\linewidth]{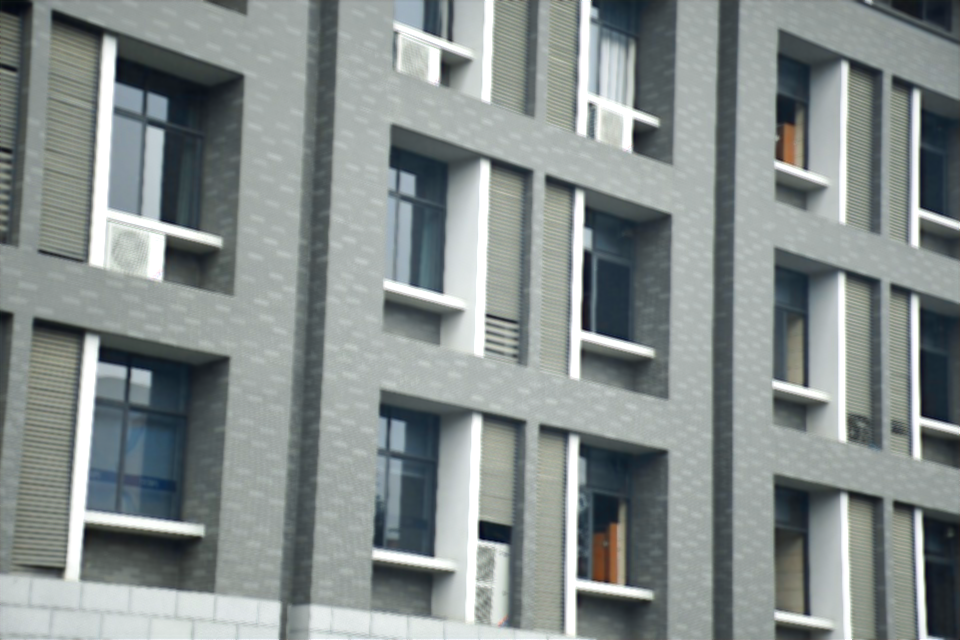}
        \captionsetup{font={footnotesize,bf,stretch=1},justification=raggedright}
        \caption{Restormer~\cite{204} }
        \label{fig:huawei_gt}
    \end{subfigure}

    \begin{subfigure}{0.48\linewidth}
        \centering
        \includegraphics[width=\linewidth]{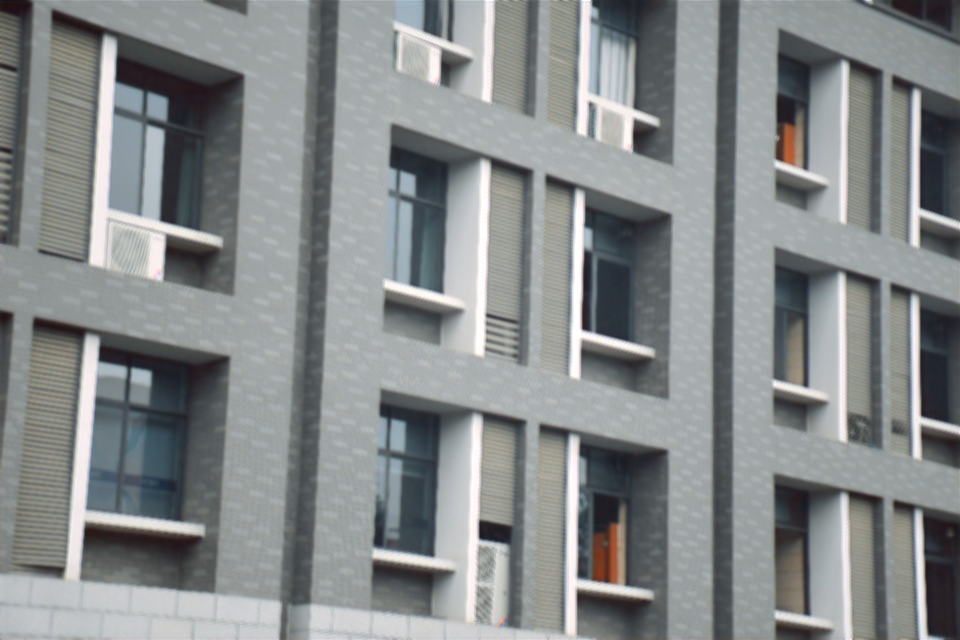}
        \captionsetup{font={footnotesize,bf,stretch=1},justification=raggedright}
        \caption{SNR-Net~\cite{inproceedings} }
        \label{fig:huawei_input}
    \end{subfigure}
    \hfill
    \begin{subfigure}{0.48\linewidth}
        \centering
        \includegraphics[width=\linewidth]{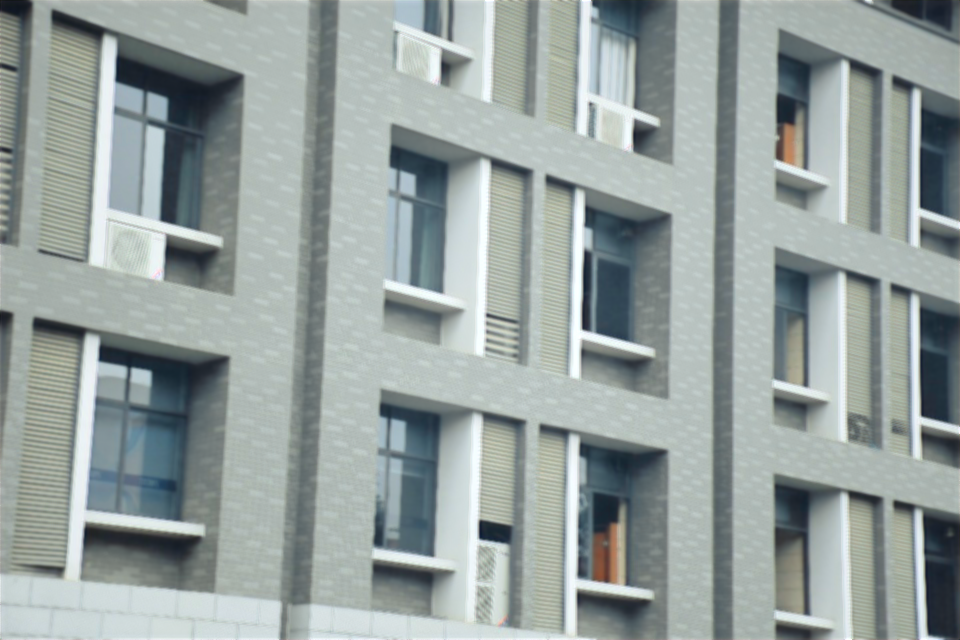}
        \captionsetup{font={footnotesize,bf,stretch=1},justification=raggedright}
        \caption{Retinexformer~\cite{205}}
        \label{fig:huawei_deepupe}
    \end{subfigure}
    
    \medskip  
    
    \begin{subfigure}{0.48\linewidth}
        \centering
        \includegraphics[width=\linewidth]{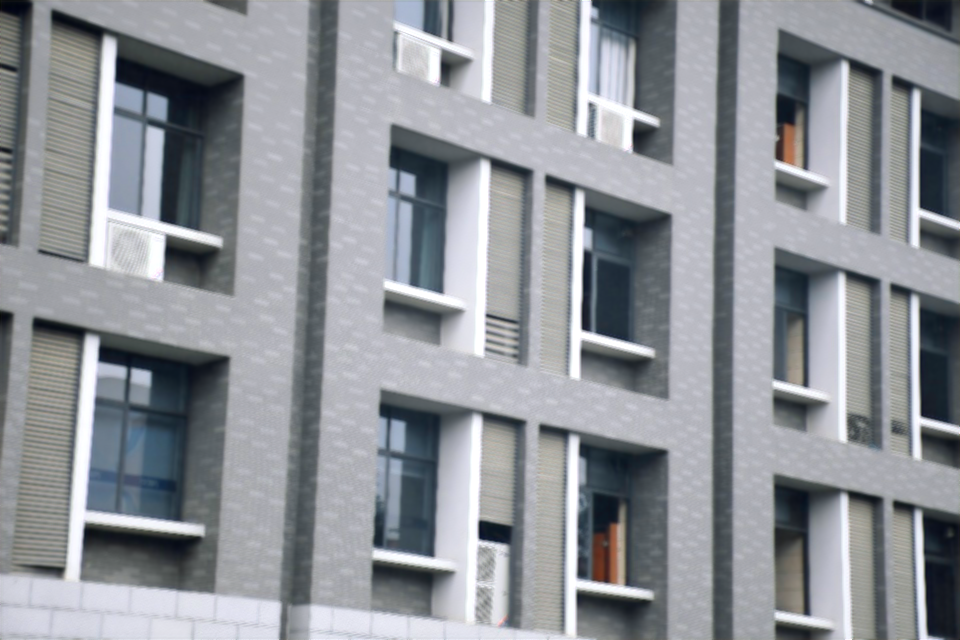}
        \captionsetup{font={footnotesize,bf,stretch=1},justification=raggedright}
        \caption{LightenDiffusion~\cite{jiang2024lightendiffusionunsupervisedlowlightimage}}
        \label{fig:huawei_ours}
    \end{subfigure}
    \hfill
    \begin{subfigure}{0.48\linewidth}
        \centering
        \includegraphics[width=\linewidth]{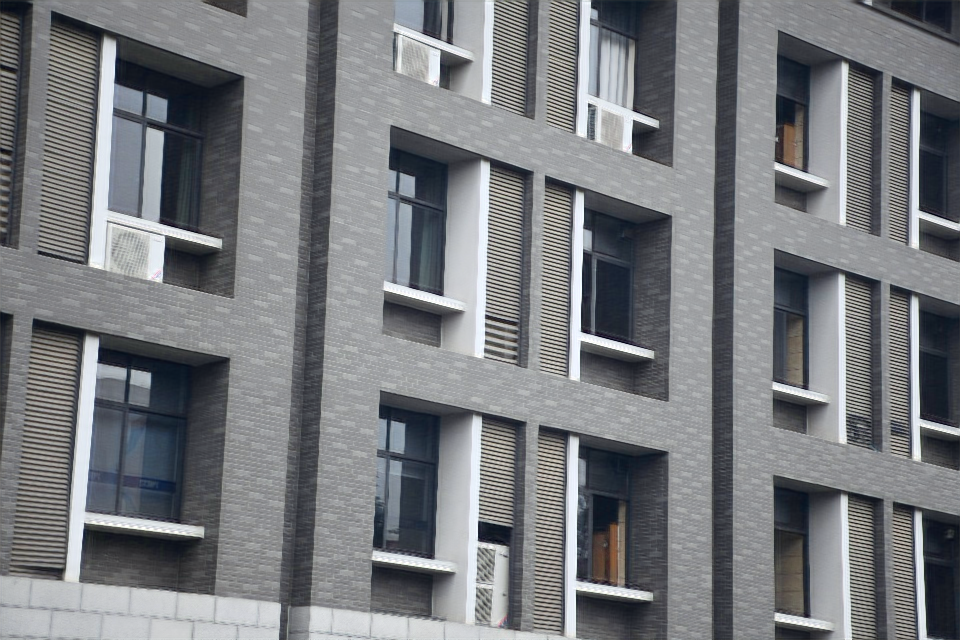}
        \captionsetup{font={footnotesize,bf,stretch=1},justification=raggedright}
        \caption{Ours(GM-MOE)}
        \label{fig:huawei_gt}
    \end{subfigure}

    \caption{\textbf{Visual comparison on the Nikon dataset.} The enhancement effect of GM-MOE is closer to the original picture than other networks.}
    \label{fig:nikon}
\end{figure}


The main text only shows a partially enlarged view of the water bottle area from the Huawei dataset. As shown in Fig.~\ref{fig:1}, we further give a full view of this photo so readers can more fully observe the results. It can be seen that for areas with patterns and text, the enhancement results of other networks are often blurry, while the text boundaries in the image processed by the GM-MOE method are clearer, the colors are more distinct, and there is no obvious blurring.

As can be seen in Fig.~\ref{fig:lol}, the network enhanced by GM-MOE does not experience color distortion or overexposure compared to other networks.

In addition, we also show a photograph of a building from the Nikon dataset, as shown in Fig.~\ref{fig:nikon},  By comparing the enhancement effects of different methods (such as DeepUPE), it can be found that the output of some methods has serious color distortion or loss of detail. In contrast, our method performs better in terms of color reproduction and detail retention, and the enhanced map is closest to the real map (Ground Truth). From a quantitative perspective, our method also achieves the highest scores in metrics such as PSNR and SSIM, which further proves its superiority.

\subsection{About the dataset}
LOL-v1~\cite{wei2018deep} is a classic low-light image enhancement dataset that covers a variety of scenes and is used to test the low-light processing effects of models in different real-world scenarios. Compared to LOL-v1~\cite{wei2018deep}, LOLv2-Real~\cite{9328179} provides more diverse lighting scenarios, while LOL-v2-Synthetifutc generates a wider range of scenarios through artificial low-light simulation algorithms, which are mainly used to evaluate the generalization ability of the model. The LSRW-Huawei~\cite{Hai_2023} and LSRW-Nikon~\cite{Hai_2023} datasets, which were captured by Huawei and Nikon devices respectively, contain images of real-world low-light scenes, which require a high level of detail in processing low-light photos.

Taking the LOLv2-Real dataset with enhanced data as an example, our method performs best in recovering the glass surface. As can be seen from the above figure, GM-MOE has better generalization ability on multiple datasets, especially in terms of detail processing, which is superior to other methods.

\subsection{Generalisation to SID Dataset}
On the SID benchmark~\cite{Chen_2018_CVPR}, GM--MoE attains 24.80~PSNR and 0.69~SSIM, demonstrating strong performance.
\subsection{Comparison with Competitive Baselines on LOL‑v1}
As shown in Tab.~\ref{tab:lolv1_results}, our method achieves a PSNR of \textbf{26.66\,dB}, an SSIM of \textbf{0.86}, and a LPIPS of \textbf{0.098}. 

\begin{table}[h]
    \centering
    \scriptsize  
    \setlength{\abovecaptionskip}{1mm}
    \setlength{\belowcaptionskip}{-2mm}
    \resizebox{1\linewidth}{!}{  
    \begin{tabular}{lccc}
        \toprule
        Methods & PSNR ($\uparrow$) & SSIM ($\uparrow$) & LPIPS ($\downarrow$) \\
        \midrule
        SCI~\cite{Ma_2022_CVPR}  & 14.78 & 0.53 & 0.392  \\
        NeRCo~\cite{Yang_2023_ICCV}  & 22.95 & 0.79 & 0.311  \\
        DiffLLE~\cite{yang2023}  & 22.24 & 0.79 & --     \\        LightenDiffusion~\cite{Jiang_2024_ECCV}  & 20.45 & 0.80 & 0.192  \\
        \textbf{Ours}     & \textbf{26.66} & \textbf{0.86} & \textbf{0.098} \\
        \bottomrule
    \end{tabular}
    }
    \caption{Quantitative comparison on the LOL-v1 dataset among different methods: SCI~\cite{Ma_2022_CVPR}, NeRCo~\cite{Yang_2023_ICCV}, DiffLLE~\cite{yang2023} , LightenDiffusion~\cite{Jiang_2024_ECCV}, and Ours.}
    \label{tab:lolv1_results}
\end{table}

\subsection{Perceptual Quality (LPIPS)}
As shown in Tab.~\ref{tab:lpips_results} , we report the LPIPS scores of different models on different benchmarks.
\begin{table}[h]
  \centering
  \scriptsize
  \setlength{\tabcolsep}{6pt}
  \renewcommand{\arraystretch}{1}
  \resizebox{1.0\linewidth}{!}{%
  \begin{tabular}{lccc}
    \toprule
    Method & LOL‑v1 & LOL‑v2‑Real & LOL‑v2‑Synthetic \\
    \midrule
    Retinexformer~\cite{205} & 0.129 & 0.171 & 0.059 \\
    LLFormer~\cite{ref28} & 0.167 & 0.211 & 0.066 \\
    GM--MoE (Ours) & \textbf{0.098} & \textbf{0.100} & \textbf{0.041} \\
    \bottomrule
  \end{tabular}}
  \caption{LPIPS ($\downarrow$) comparison across three LOL datasets. Lower is better.}
  \label{tab:lpips_results}
\end{table}

\subsection{How are expert modules coordinated?}
\begin{figure}[h!]         
  \centering
  \includegraphics[width=1.1\linewidth]{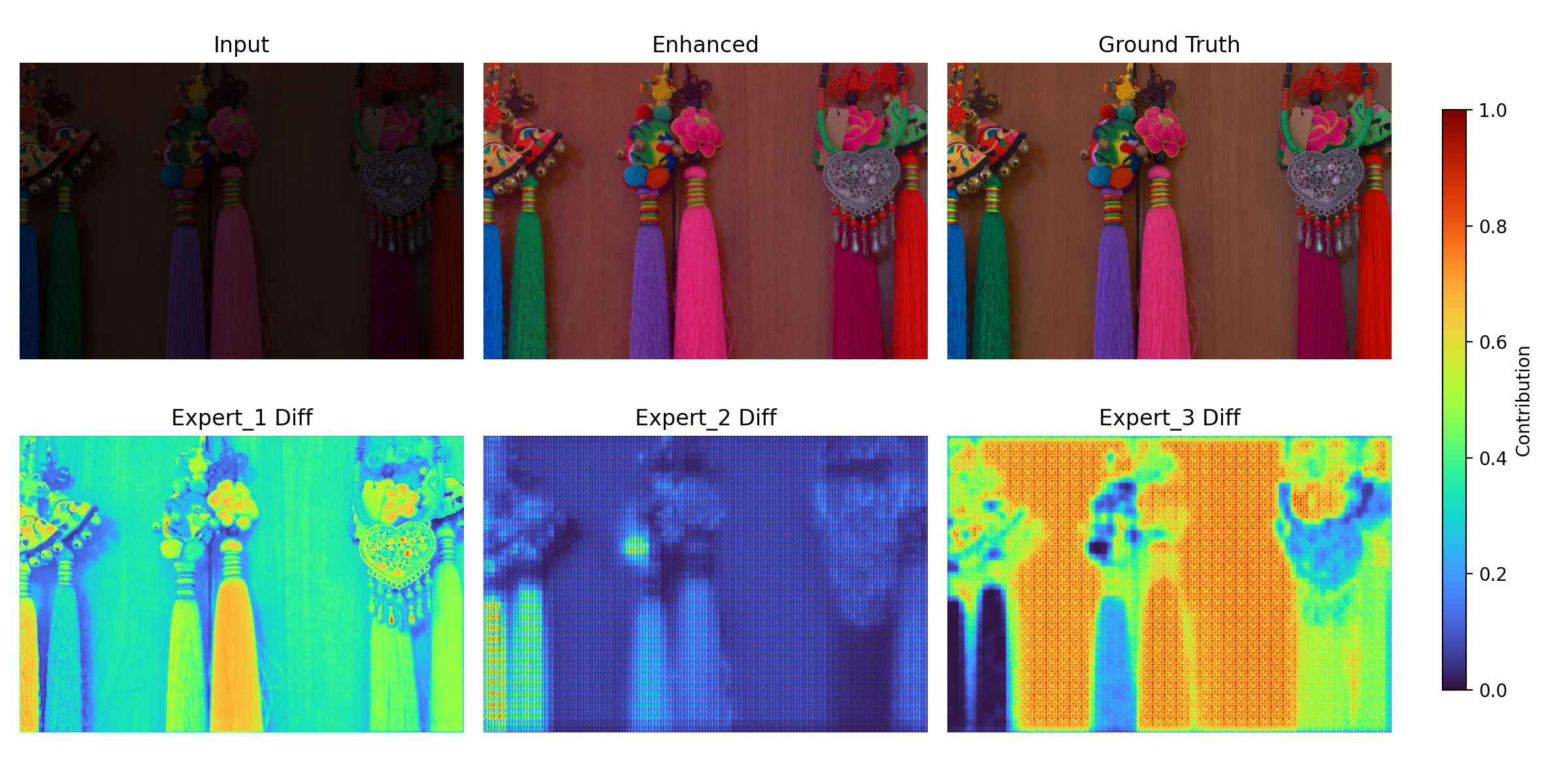}
  \caption{The heatmaps show distinct focus areas across the three experts, forming a coherent, synergistic attention distribution.}
  \label{hot} 
\end{figure}

As shown in Fig.~\ref{hot} , we enforce the use of only one expert at
a time. The heatmap of Expert1 exhibits yellow-to-red contributions in fringe details, indicating its primary role in local chroma restoration. Expert2's heatmap shows predominantly dark blue regions with only faint highlights in limited detailed areas, suggesting its specialization in fine texture recovery. Expert3 demonstrates extensive orange-yellow coverage across both the main fringe and background regions. These three heatmaps reveal spatial complementarity among the experts, enabling the final enhanced results to achieve both rich detail preservation and more realistic color reproduction.

\subsection{Computational Efficiency}
Our method delivers superior low‑light enhancement quality while incurring a computational cost of 27.2~GFLOPs, which is still practical for real‑time deployment. Future work will focus on further reducing latency and memory footprint without sacrificing restoration accuracy.

\subsection{Ablation study on network structure and expert interactions}
To explore in depth the relationship between the number of parameters and performance, we conducted three sets of experiments: fixed weights, cascaded networks, and constrained parameter growth.

\begin{table}[h]
    \centering
    \small
    \setlength{\abovecaptionskip}{1mm}
    \setlength{\belowcaptionskip}{-2mm}
    \resizebox{\linewidth}{!}{%
    \begin{tabular}{r l c c c}
        \toprule
        No. & Variant & Params (M) & PSNR ($\uparrow$) & SSIM ($\uparrow$) \\
        \midrule
        1 & Original GM-MoE & 19.99 & 23.65 & 0.80 \\
        2 & Without dynamic gating (fixed weights) & 19.86 & 21.70 & 0.71 \\
        3 & Serializing three expert networks & 19.86 & 21.34 & 0.83 \\
        4 & Experts' channels concat + $1\!\times\!1$ fusion & 20.60 & 17.84 & 0.70 \\
        \bottomrule
    \end{tabular}}
    \caption{Ablation study on network structure and expert interactions evaluated on the LOLv2-Real dataset.}
    \label{tab:ablation}
\end{table}

\subsection{Limitation and future works }
\noindent\textbf{Increase in Sub-Expert Networks and Its Impact on Performance.} Increasing the number of sub-expert networks may improve the model's performance, but it also introduces additional computational complexity. In GM-MoE, the role of the sub-expert networks is to tackle different low-light enhancement tasks, allowing the model to process various image features more specifically. Each sub-expert network focuses on different aspects of low-light image enhancement, which can lead to better performance, particularly when the tasks are well-defined and complementary.

\textbf{Scalability to Downstream Tasks.}
Currently, we have applied GM-MoE to enhance low-light images and used these enhanced images for object detection. However, future work should explore extending the GM-MoE framework to other downstream tasks. For example, video enhancement processing is a promising avenue for application. The framework may also be applicable to other tasks such as image segmentation or visual recognition, which could further demonstrate the versatility of GM-MoE.

\textbf{Adjustability of Loss Functions and Their Impact on the Model.}
Currently, we use PSNR as the primary metric for image quality. However, in future work, we should investigate how adjusting the loss function impacts the overall performance of the model. Experimenting with alternative loss functions, such as perceptual loss or adversarial loss, may provide better results in preserving image details and enhancing visual quality, especially for more complex tasks. The choice of loss function can significantly affect the model's ability to generalize across different datasets and tasks.

In future work, we will further explore the application of the GM-MoE model. First, we will study how to improve the computational efficiency of the model. Second, we will explore the use of GM-MoE-enhanced images in downstream tasks such as image segmentation and video enhancement to verify its versatility and adaptability.
\section*{Acknowledgments} This work was supported by the Xinjiang Uygur Autonomous Region Tianshan Ying Talents Leading Talents Program for Scientific and Technological Innovation (2023TSYCLJ0025), the National Natural Science Foundation of China (No. 62266044, 61563052, 61862061), and the National Key Research and Development Program of China (2021YFB2802100).

{
    \small
    \bibliographystyle{ieeenat_fullname}
    \bibliography{main}
}


\end{document}